\documentclass{egpubl}
\usepackage{VMV2025}

\usepackage{subcaption}
\usepackage{float}
\usepackage[inline]{enumitem}
\usepackage{amsmath}
\usepackage{bm}
\usepackage{amssymb}
\usepackage[a4paper]{geometry}
\usepackage[pdftex]{graphicx}
 
 \WsPaper           

\usepackage[T1]{fontenc}
\usepackage{times}
\usepackage{newtxtext,newtxmath}

\usepackage{cite}  
\BibtexOrBiblatex
\electronicVersion
\PrintedOrElectronic
\ifpdf \usepackage[pdftex]{graphicx} \pdfcompresslevel=9
\else \usepackage[dvips]{graphicx} \fi

\usepackage{egweblnk} 

\title[Neural Acquisition \& Representation of Subsurface Scattering]%
      {Neural Acquisition \& Representation of Subsurface Scattering}

\author[A. Majumdar, R. Braun, H. Lensch]{%
  Arjun Majumdar$^{1}$
  Raphael Braun$^{1}$
  Hendrik Lensch$^{1}$ \\
  $^1$University of Tübingen, Germany \\
}

\usepackage[table]{xcolor} 
\definecolor{turquoise}{cmyk}{0.65,0,0.1,0.3}
\definecolor{purple}{rgb}{0.65,0,0.65}
\definecolor{dark_green}{rgb}{0, 0.5, 0}
\definecolor{orange}{rgb}{0.8, 0.6, 0.2}
\definecolor{red}{rgb}{0.8, 0.2, 0.2}
\definecolor{darkred}{rgb}{0.6, 0.1, 0.05}
\definecolor{blueish}{rgb}{0.0, 0.3, .6}
\definecolor{light_gray}{rgb}{0.7, 0.7, .7}
\definecolor{pink}{rgb}{1, 0, 1}
\definecolor{greyblue}{rgb}{0.25, 0.25, 1}

\usepackage{tikz}
\usepackage{pgfplots}
\pgfplotsset{compat=1.18} 
\definecolor{MyBlue}{rgb}{0,0,1}
\definecolor{MyCyan}{rgb}{0,1,1}
\definecolor{MyGreen}{rgb}{0,1,0}
\definecolor{MyYellow}{rgb}{1,1,0}
\definecolor{MyRed}{rgb}{1,0,0}

\begin{document}


\maketitle
\begin{abstract}
We present a method to acquire and estimate the sub-surface scattering properties of light transport at a highly detailed level by learning the pixel footprint response at each point on the object surface. The reconstruction leverages 3D scanning techniques as input to a U-Net CNN. A stereo projector-camera setup using phase-shifted profilometry (PSP) patterns efficiently captures the data for a variety of scattering objects. Reconstructing dense pixel footprints allows for relighting with arbitrary high-resolution projector patterns.  The final output is a relit color image. Qualitative and quantitative comparison against illuminated real-world captured images demonstrate that the predicted footprints are almost identical to the actual responses.  The same model is trained for multiple views across multiple objects such that the learned representations can be used to generalize to unseen sub-surface scattering materials as well.
   
   



\printccsdesc   
\end{abstract}  
\section{Introduction}
In light transport \emph{subsurface scattering} refers to the phenomena of light
entering a surface, being scattered inside the material and leaving it at
another location than where it entered. All non metal materials allow light to
enter to various depths, which depends on the individual material. In case of
glass like materials light passes through without being scattered. On the
other side there are materials like wood, paper or concrete, which are
considered opaque, as the light only penetrates to microscopic depths before
being scattered back. Those extreme cases can be easily described with analytic
BSSRDFs, for glass a clear dielectric model and for opaque materials a diffuse model. However in between there are materials such as wax, marble,
organic objects, fruits, leaves, flowers, skin and others, which are not as
straight forward. There are analytic models, that describe light transport
inside the material as diffusion process \cite{practical_sss_light_transport},
which in cases of geometry in form of semi-infinite slabs have exact solutions.
Recently deep learning based approaches appeared which push the idea of closed
form solutions to arbitrary curved surfaces \cite{learned_sss_vicini} while
maintaining the efficiency of surface based models.
Alternatively it is always possible to simulate the volumetric light transport
inside a material via path tracing~\cite{manuka, renderman}, which however
requires a detailed volumetric representation of the optical properties inside
an object to get correct results for complex materials such as alabaster or any
object with complex volumetric composition. Creating or measuring such a
volumetric representation is challenging, as only
the scattering result on the surface are directly observable, not the underlying
density fluctuations and phase function changes. Instead of describing the
complex light transport using surface and volume representation we use an image
based rendering approach, where the impact of every light ray on the scene is
captured as image. Here relighting an object just means adding together the
correct images with the correct scaling factors. In practice it is however very costly
to capture such an image dataset. The non-local nature of
Subsurface Scattering requires the acquisition of thousands of images, even when the effect of multiple rays are captured in a single picture.

In this paper, we propose a novel, completely data-driven method for image-based modeling of subsurface scattering objects over multiple views and across multiple objects, which after training can predict the required images for relighting without the need for excessive acquisitions.
Instead, high-frequency phase-shifted sinusoidal patterns are used as input to a U-Net-based Convolutional Neural Network (CNN) \cite{unet_paper}. 
These patterns are the same patterns that are used to capture precise 3D geometry of the object during 3D scanning. 
Our U-Net predicts a per pixel anisotropic footprint in image space expressing the subsurface scattering response to incident illumination at that point on the surface. 
It learns to generalize to unseen views and objects.
Furthermore, our method can be trained from images of a single object and,
unlike other methods, does not require to learn from many other objects, nor is it probabilistically generative in nature. As a result, it will not suffer from prior distribution bias. Also, since it consists of a single U-Net network, training it is relatively cheap, efficient and easy.

We perform comprehensive qualitative and quantitative evaluations of our learned footprint responses to render the object under various projection patterns and compare the resulting images with the camera captured ground truth to demonstrate the performance of our model on objects with challenging material properties and geometry. Since the focus of our method is on learning and estimating the footprint response of each surface pixel, we perform extensive evaluations on relit results. 

\noindent In summary, we claim the following contributions:
\begin{itemize}
    \item We learn anisotropic pixel footprint responses to capture subsurface
    scattering properties by only using projected high-frequency phase-shift
    patterns which can at the same time be used for 3D scanning.
    \item A neural network estimates footprint responses for each point on the
    object  which then are used to perform relighting with spatially varying
    light patterns in RGB colorspace of the image.
    \item Our neural network architecture is able to generalize across multiple
    views and multiple objects, which we verify both qualitatively and
    quantitatively.
\end{itemize}

We compute image based sub-surface scattering (SSS) estimation. The actual material parameters for modeling the BSSRDF is not inferred. We only learn anisotropic SSS footprint responses which encodes both the geometry and the material properties and can be used for image based relighting.

\section{Related Works}

The acquisition of material properties has a long-standing tradition, including the measurement of subsurface scattering parameters or general relightable representations, e.g.\ reflectance fields~\cite{debevec2000acquiring}. 
While initial works sampled directly the impulse response to incident illumination, fixed wavelet noise patterns~\cite{peers2003wavelet}, compressive sensing~\cite{peers2009compressive} or adaptive acquisition schemes~\cite{sen2005dual,garg2006symmetric,o2010optical,o2012primal} exploiting the directionally duality of the light transport have been employed for faster acquisition of general light transport matrices. 
The required number of acquisition patterns still remains substantial. 
After some initial training over multiple objects, any new object and new view in our approach only requires about six input images to recover the spatially varying pixel footprint for subsurface scattering. 

After the introduction of the first practical BSSRDF model~\cite{jensen2001practical} the specific estimation of subsurface scattering parameters has been performed using dedicated point-wise illumination~\cite{jensen2001practical,weyrich2006analysis},
For entire objects, Goesele et al.\ \cite{goesele2004disco} have presented a system, where a single laser beam scans every surface point to measure the resulting camera footprint, requiring thousands of individual measurements.

A few neural network-based relighting approaches specifically focus on relighting of subsurface scattering objects~\cite{zheng2021neural,lyu2022neural,yu2023learning,zhu2023neural,dihlmannsubsurface} from a smaller set of images. 
However, those approaches assume distant illumination and often use one-light-source-at-a-time (OLAT) as training data.


Structured light scanning patterns, specifically high-frequency patterns, have been used to separate the observed reflected radiance wrt.\ being caused by local or global illumination~\cite{nayar2006fast}. 
There are existing methods that leverage high-frequency PSP for sub-surface scattering materials in 3D scanning: \cite{fuchs2008combining,mod_phase_shift_3d,polarization_phaseshift,micro_phaseshift,embed_phaseshift}, etc. However, they only focus on removing the unwanted effect of subsurface scattering on the depth estimation rather than measuring and representing the subsurface scattering per se for relighting. 
Geiger et al.\ \cite{improved_topo_sss} utilize structured light projection to get correct topography for scattering objects as light undergoing volume scattering inside the object results in erroneous outputs. These errors are studied using  Monte Carlo simulations, together with an additional method to correct the errors by quantifying the light propagation. 
Other approaches, e.g.\ ~\cite{kienle1996spatially,kikuchi2023development}, explicitly exploit the properties of high-frequency binary patterns and polarization filters to directly measure local isotropic scattering parameters of human skin.

To the best of our knowledge, we for the first time use high-frequency PSP images as the only input to reconstruct spatially-varying, potentially anisotropic scattering footprints by a neural network to obtain a more accurate relightable model.

A neural architecture for efficiently sampling Bidirectional Scattering Surface Reflectance Distribution Functions (BSSRDF) on complex 3D surfaces is proposed by Vicini et al.\ \cite{learned_sss_vicini}. The system includes three networks: 1) A Conditional VAE (CVAE), trained on outputs from a slow but accurate volumetric path tracer, learns to generate similar scattering samples much faster by conditioning on input parameters. 2) A scale factor regression MLP adjusts the CVAE output to account for material-dependent absorption differences. 3) A preprocessing feature network transforms raw inputs into a learned feature space, enabling more effective conditioning for the CVAE and MLP.

Our method, by comparison, does not depend on any pre-trained VAE method and learns the anisotropic footprint responses directly from the input images. Since it does not depend on multiple networks and just consists of a single U-Net CNN architecture, training is more efficient and easier. 
VAEs model complex data distributions by maximizing the evidence lower bound (ELBO) loss and minimizing the reconstruction loss \cite{embrace_the_gap}. They might add bias due to the noisy, uninformative prior that is used in KL-loss computation, whereas our method learns the actual latent representation, thereby remaining bias-free. Lastly, VAEs might also suffer from posterior collapse \cite{vae_posterior_collapse,regressing_vae_posterior_collapse,dont_blame_elbo}, which our CNN does not. Our primary focus is to learn from the camera-captured ground truth anisotropic footprint responses as faithfully as possible.


\section{Background}
Our acquisition of subsurface scattering representations is based on N-step phase-shifted profilometry (PSP) which typically is applied to perform accurate 3D scanning. 


Assuming a camera/projector setup with a linearized projector, a set of virtual sinusoidal patterns can be created and then projected onto the object in the scene, captured by the corresponding offset camera.
These virtual sinusoidal patterns in the projector space can be represented as:
\begin{equation}
I^p(x^p, y^p) = a^p + b^p cos(2\pi f_0^p x^p + \delta_n),
\end{equation}
where $(x_p, y_p)$ represents the projector pixel coordinate, $a^p$ depicts the direct component of intensity, $b^p$ represents the amplitude and $f_0^p$ is the frequency of the sinusoidal fringe in period per pixel. $\delta_n$ is the phase-shift index and can be represented as $\delta_n = \frac{2 \pi}{N}n$ for the standard N-step PSP \cite{psp_review_paper}. 

The resulting camera images can then be expressed as:
\begin{equation}
    I_n(x,y) = A(x,y) + B(x,y) cos \left(\phi(x,y) - \frac{2 \pi}{N} \right)
\end{equation}

where, $A(x,y)$ is the average intensity for the pattern brightness and background illumination, $B(x,y)$ is the intensity modulation pertaining to pattern contrast and surface reflectivity and $n$ is the phase-shift index ranging from $n=0,1,...N$.

The wrapped phase map can be computed by using the equation for each point (x,y) in camera or image space:
\begin{equation}
    \phi(x, y) = \tan^{-1} \frac{\sum_{n=0}^{N-1} I_n(x, y) \sin(2\pi n/N)}{\sum_{n=0}^{N-1} I_n(x, y) \cos(2\pi n/N)}
\end{equation}

The wrapped phase $\phi(x,y)$ is in the range $[-\pi, \pi]$. Since there are three unknowns: $A(x,y)$, $B(x,y)$ and $\phi(x,y)$, we usually require three images to compute them \cite{psp_review_paper,tpu_review_paper}. To convert the wrapped phase to unwrapped or continuous phase map, we use the equation:
\begin{equation}
    \Phi(x,y) = \phi(x,y) + 2\pi k(x,y),
\end{equation}

where $k(x,y)$ is the fringe integer number that indicates the fringe orders. The majority of the phase-unwrapping algorithms deal with computing this fringe integer number as correctly, quickly and faithfully as possible \cite{tpu_review_paper,psp_review_paper}.
The unwrapped phase directly correlates to the disparity between the camera and the projector in a stereo setup and therefore is related to depth.

Broadly, PSP is divided into spatial and temporal phase unwrapping (TPU), where in the former, neighboring pixels' information is used to unwrap a pixel, whereas in the latter, only time-varying information is used to unwrap a pixel. Consequently, temporal phase unwrapping is more robust and is the preferred way as it has a better ability to measure surface discontinuities \cite{tpu_comparative_study}.

However, for challenging objects, especially for subsurface scattering objects, traditional PSP TPU is insufficient due to the global effects of subsurface scattering disturbing the acquired lower unit frequency phase maps, which are required to unwrap the higher frequencies \cite{mod_phase_shift_3d,polarization_phaseshift}. 
An alternate approach for such challenging materials and complex geometrical objects employs interferometry-based hierarchical phase-unwrapping. 
Instead of projecting and capturing a unit-frequency, they use the principles of interferometry wherein a synthetic wavelength is used for unwrapping the highest-frequency phase map, which is free from artifacts due to global effects such as subsurface scattering. 
Additionally, high-frequency sinusoidal patterns can also be leveraged to decompose a captured scene into its direct and global components~\cite{ld_lg_seperation}. 
Here, the recovered amplitude $B$ is directly related to the local component due direct illumination turned on or off, while the recovered average $A$ corresponds to the global component representing all global illumination effects such as ambient illumination, interreflections and, relevant for us, subsurface scattering. 

U-Net is a Convolutional Neural Network (CNN) which was originally proposed for medical domain based image segmentation task by outputting a mask for the given input images \cite{unet_paper}.



\section{Method}
\begin{figure}[ht]
    \centering
    \includegraphics[scale=0.26]{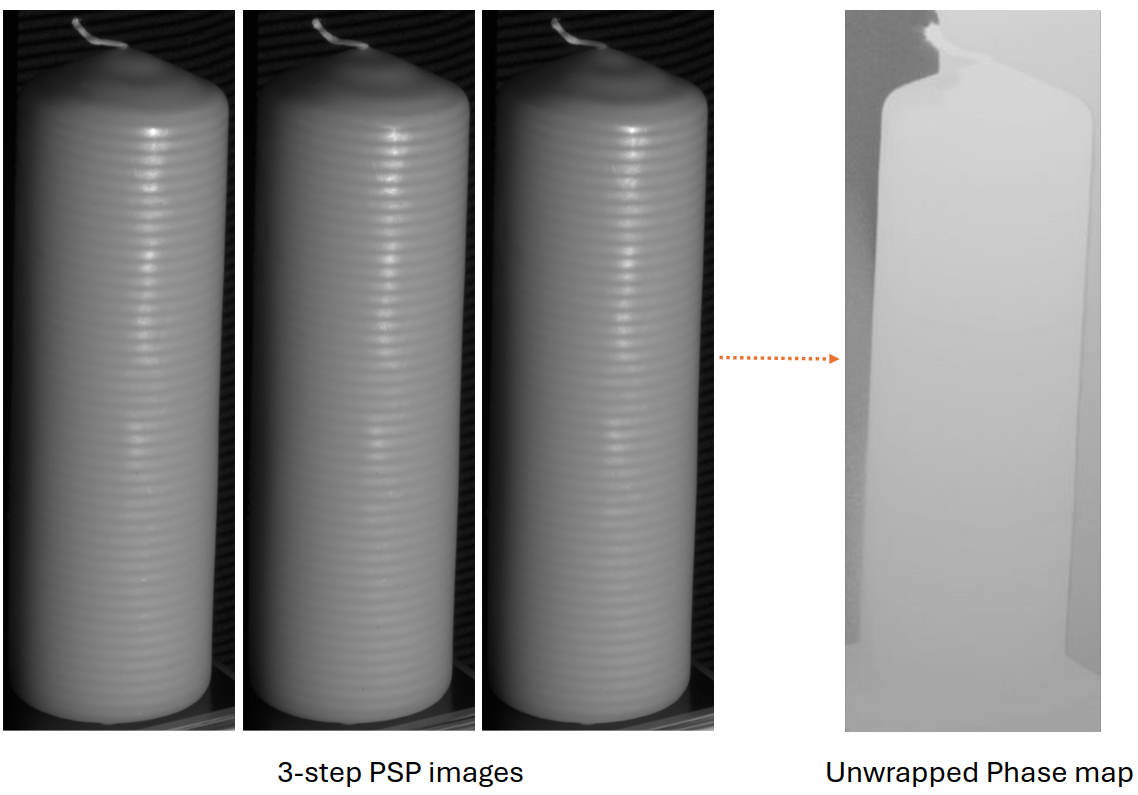}
    \caption{3-step phase-shifted proliferometry (PSP) example. The same pattern can be used to measure disparity as well as for recovering spatially varying scattering footprints. }
    \label{fig:psp_example}
\end{figure}

Our image and learning-based model of subsurface scattering assumes the
additive nature of light. We estimate the anisotropic pixel footprint response
at each point of a subsurface scattering object by using a set of high-frequency
(4 pixels per cycle) PSP images. We employ a camera/projector stereo setup. A
BenQ X1300I 3D DLP projector of 1080x1920 pixel resolution projects different
patterns onto the subsurface object in the scene, which is then captured by a
FLIR Oryx 10GigE camera of 3000x3200 resolution. 
Fig.~\ref{fig:psp_example} shows an example of a scattering candle: for 3-step PSP, three phase-shifted horizontal patterns with a phase-shift of $\frac{2\pi}{3}$  are visualized together with their corresponding unwrapped phase map. 
We use this 3D scanned output to obtain a pixel-wise mapping for each camera pixel to its corresponding projector pixel. This correspondence can be used for 3D reconstruction and is used during relighting to know which projector pixel illuminates which camera coordinate directly.

\begin{figure}[ht]
    \centering
    \includegraphics[scale=0.45]{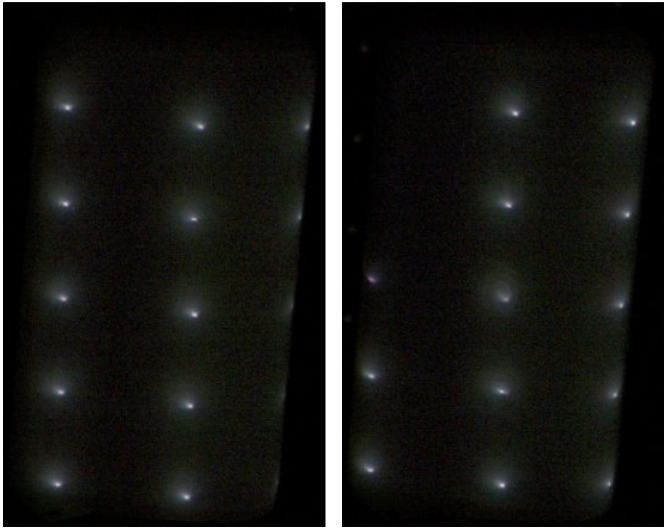}
    \caption{Camera captured target ground truth. As training data we acquire densely sampled scattering footprints by illuminating individual dots with a shifting grid.  }
    \label{fig:dots_example_pic}
\end{figure}

For one view of an object we capture the pixel responses due to subsurface scattering explicitly and use them as ground truth for training the U-net. 
We capture the subsurface scattering-based pixel footprint responses by
projecting a grid of dots $(m,m)$ onto the object. Here, $m$ denotes the
distance in pixels in both the x and y axes in the projector space.  Capturing
each image with just one projected dot onto the surface of the object is clearly
infeasible due to time and space complexity reasons. Therefore, we parallelize
the capture of multiple surface scattering points by projecting multiple dots
onto the object and then capturing all of them in a single camera image. The
total number of acquired images is therefore $m \times m$. For full ground truth
acquisition, we shift the grid by 1 pixel such that in the end, we have covered
the entire object's surface area at least once. 
The pixel distance $m$ needs to be chosen such that the observed footprints clearly do not overlap. If $m$ is small, i.e., 40 pixels or less, then the light from the neighboring adjacent coordinates might affect the response for the current coordinate and thereby lead to a biased captured response. We use $(55,55)$ as standard for all our acquisitions. An example acquisition for a bar of soap is shown in Fig.\ \ref{fig:dots_example_pic}. The two images depict the shifted projected and captured patterns.

The camera captured images are in raw, mosaiced, luminance-only data format. We apply the following data pre-processing steps on them before using them for our deep learning training:
\begin{enumerate*}[label=(\roman*)]
    \item Median filtering: we apply channel-wise filtering for $Red$, $Green_1$, $Green_2$ and $Blue$ channels to get denoised outputs
    \item Demosaicing: \cite{malvar_demosaicing}, \cite{menon_demosaicing}, \cite{linear_image_demosaicing} algorithms are used to obtain demosaiced RGB colored images.
    \item We use Segment Anything Model \cite{sam_paper} to get a mask of the object while excluding the background so that only footprint responses on the object are used for training the U-Net and no other non-scattering response can harm the learned representations
    \item high-frequency PSP input images are captured as input for U-Net, which never sees the target dots
\end{enumerate*}

The processed data is split into train-test sets at a 85:15 ratio. 

\subsection{Learning Footprint Prediction}

\begin{figure}[ht]
    \centering
    \includegraphics[scale=0.4]{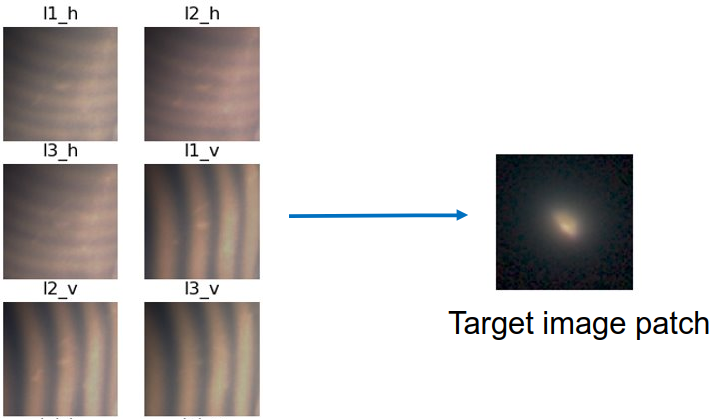}
    \caption{Attention U-Net data flow. Given horizontal and vertical PSP pattern around camera pixel $x,y$ as input, the U-Net predicts the corresponding anisotropic footprint.}
    \label{fig:sine2dotmap}
\end{figure}

For estimating the pixel footprints from the horizontal and vertical high-frequency PSP input we use an Attention U-Net architecture \cite{unet_paper,unet_attention}. It learns to output the anisotropic footprint response at that point as schematically shown in Figure~\ref{fig:sine2dotmap}.
Please note that during inference, the U-Net never sees any of the footprint response. It learns to estimate these only from the high-frequency PSP input patterns at each point of the object.

A vanilla U-Net was not able to faithfully learn the anisotropic footprint responses as the anisotropy seems to vary for each surface point on the object for each view and across objects. Therefore, we shifted to an Attention U-Net variant \cite{unet_attention} where the attention gate allows the U-Net to focus on targets of varying shapes and sizes. For all of our experiments, we use the Attention U-Net variant only. Details about the deep learning training is provided in the supplementary.

Due to the extreme distribution in the footprints with very few rather bright spots, some anisotropic decay and otherwise many rather small values which are overlapping with the camera's noise floor, it is a challenge to determine a proper loss function. 

Initially the Attention U-Net was trained with a Mean-Squared Error (MSE) cost function:
\begin{equation}
    \text{MSE} = \frac{1}{n} \sum_{i=1}^{n} (y_i - \hat{y}_i)^2 
\end{equation}
Since the majority of each input patch is dark and thereby the resulting output of the network might also be proportionately small, the MSE often produces very small values. 
As a result, the gradient signal might be insufficient to update the parameters of the network. Trying to alleviate this, we also experimented with Mean Squared Logarithmic Error (MSLE) cost function:
\begin{equation}
    \text{MSLE} = \frac{1}{n} \sum_{i=1}^{n} (log(y_i) - log(\hat{y}_i))^2 
\end{equation}

This log transformation leads to good loss values and proper gradient signals, but it truncates the smooth anisotropic falloff as it punishes small deviations heavily. Consequently, the learned footprint responses don't produce the smooth blurring visualization effects, which is a typical characteristic of sub-surface scattering.


Finally, we settled with an inverse-Gaussian weighted MSE cost function with the following weighting scheme:
$P' = P + \epsilon$, 
$W = w_{\max} \cdot \exp\left(-\frac{(\mathbf{P}')^2}{2\sigma^2}\right)$ and 
$W = \text{clamp}(\mathbf{W}, 1.0, w_{\max})$
where $\mathbf{P} \in \mathbb{R}^{B \times C \times H \times W}$ is the target subsurface scattering footprint, $\sigma = 0.1$ controls the decay rate of the weighting function, $\epsilon = 10^{-8}$ is a small numerical shift to avoid division by zero, $w_{\max} = 100.0$ is the maximum weight value, and $\mathbf{W}$ are the resulting inverse Gaussian weights that assign higher values to pixels with smaller target values.

The weighted MSE cost function used for all of our trainings are:
\begin{equation}
    \text{inv wtd Gauss MSE} = \frac{1}{n} \sum_{i=1}^{n} (\mathbf{W_i} \cdot y_i - \mathbf{W_i} \cdot  \hat{y}_i)^2 
\end{equation}

The Attention U-Net is simultaneously trained on the footprints of a couple of different objects, typically providing one or two views of the pixel-grid pattern of Figure~\ref{fig:dots_example_pic}. The resulting network, however, generalizes to other views and other objects with different material properties. 

In total, we train the network for 150 epochs on roughly 29k pairs of sinusoidal tiles and footprints in total. Although this might vary depending on the size of the object. We use Adam (lr = $1 \times 10^{-4}$) as our gradient-descent optimizer. We monitor both train and test losses for the train-test splits that help in monitoring the progress of the neural network training, diagnostics and troubleshooting.


\subsection{Inference and Relighting}
\label{sec:inference}
For a novel view of an object, n-step PSP patterns are acquired with in total 54 images. Only the six highest frequency images (horizontal and vertical with three phase shifts each) will be used for the footprint prediction, the other 48 images are just used for hierarchical temporal phase unwrapping for phase-shift profilometry fringe pattern projections.

We remove the background that did not receive sufficient projector illumination. For all remaining camera pixels, the U-Net predicts a separate footprint by cropping the corresponding input tiles around the pixel. 

For relighting, the unwrapped camera-projector pixel correspondence map is used to identify which projector pixel (with subpixel accuracy) illuminates the corresponding camera pixel. The bilinearly interpolated project pixel's color $p(x,y)$ will be used as a color weight to multiply the camera pixel's footprint. Finally, the weighted footprint is splatted into the final framebuffer until all camera pixels are processed. 

We perform white balance color correction by projecting and capturing pure red,
green and blue images and computing a color correction matrix using Least
Squares/Pseudo-inverse method against a reconstructed red, green and blue relit image.
The camera captured pure red, green, blue and white images have
known colors which are normalized to a maximum intensity of 1.0. We compute a
white balance transformation matrix by solving a system of least squares where
$I_{\text{rec}} \cdot \mathbf{X} = I_{\text{cam}}$ where the pixels of the reconstructed red,
green and blue images in the vector $I_{\text{rec}}$ times the unknown
color correction matrix $\mathbf{X}$ equals the captured pixel colors
$I_{\text{cam}}$. This transformation matrix $\mathbf{X}$ is later multiplied to
every reconstructed output images to obtain color corrected results. During
acquisition we ensure that the cameras white balance settings stay constant to ensure the transferability of $\mathbf{X}$ between acquisitions.

\section{Results}
To demonstrate the performance of our approach we investigate both the quality of the estimated footprint as well as the resulting relit output. 

\begin{table}[htbp]
\centering
\small
\begin{tabular}{|l|c|c|c|c|c|}
\hline
\textbf{Object+View} & \textbf{Dataset} & \textbf{MSE$^*$} & \textbf{PSNR} & \textbf{SSIM} & \textbf{LPIPS$^*$} \\
\hline
Soap-front & Train & $4.07$ & 29.59 & 0.997 & $6.33$ \\
\hline
Soap-front & Test & $4.09$ & 29.59 & 0.997 & $5.99$ \\
\hline
Soap-back & Train & $3.89$ & 27.43 & 0.997 & $3.95$ \\
\hline
Soap-back & Test & $3.81$ & 27.48 & 0.997 & $3.73$ \\
\hline
Orange-front & Train & $4.09$ & 28.89 & 0.996 & $3.76$ \\
\hline
Orange-front & Test & $4.83$ & 28.85 & 0.996 & $3.45$ \\
\hline
Orange-back & Train & $3.6$ & 33.4 & 0.998 & $4.85$ \\
\hline
Orange-back & Test & $3.4$ & 33.41 & 0.998 & $4.14$ \\
\hline
Leaf-front & Train & $3.75$ & 23.82 & 0.996 & $0.93$ \\
\hline
Leaf-front & Test & $3.78$ & 23.76 & 0.996 & $0.91$ \\
\hline
Leaf-back & Train & $6.25$ & 26.91 & 0.996 & $14.4$ \\
\hline
Leaf-back & Test & $6.01$ & 26.93 & 0.996 & $13$ \\
\hline
\end{tabular}
\caption{Quantitative metrics on footprint reconstruction. $^*$(MSE$\times10^{-7}$, LPIPS$\times10^{-4}$). }
\label{tab:patch_comparisons}
\end{table}

\begin{figure}[htb]
\centering
\begin{tabular}{ccc}
\includegraphics[width=0.12\textwidth]{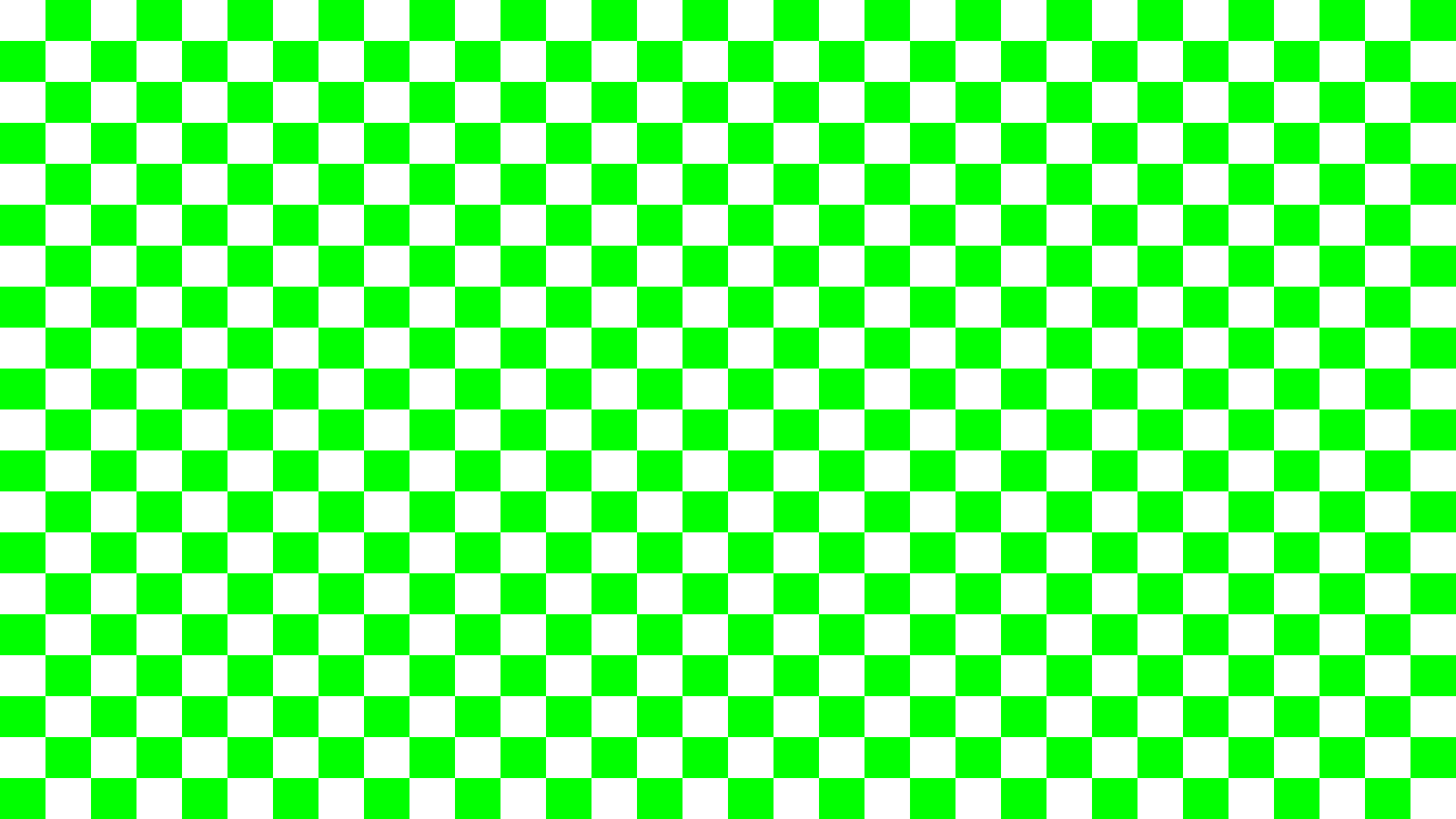} &
\includegraphics[width=0.12\textwidth]{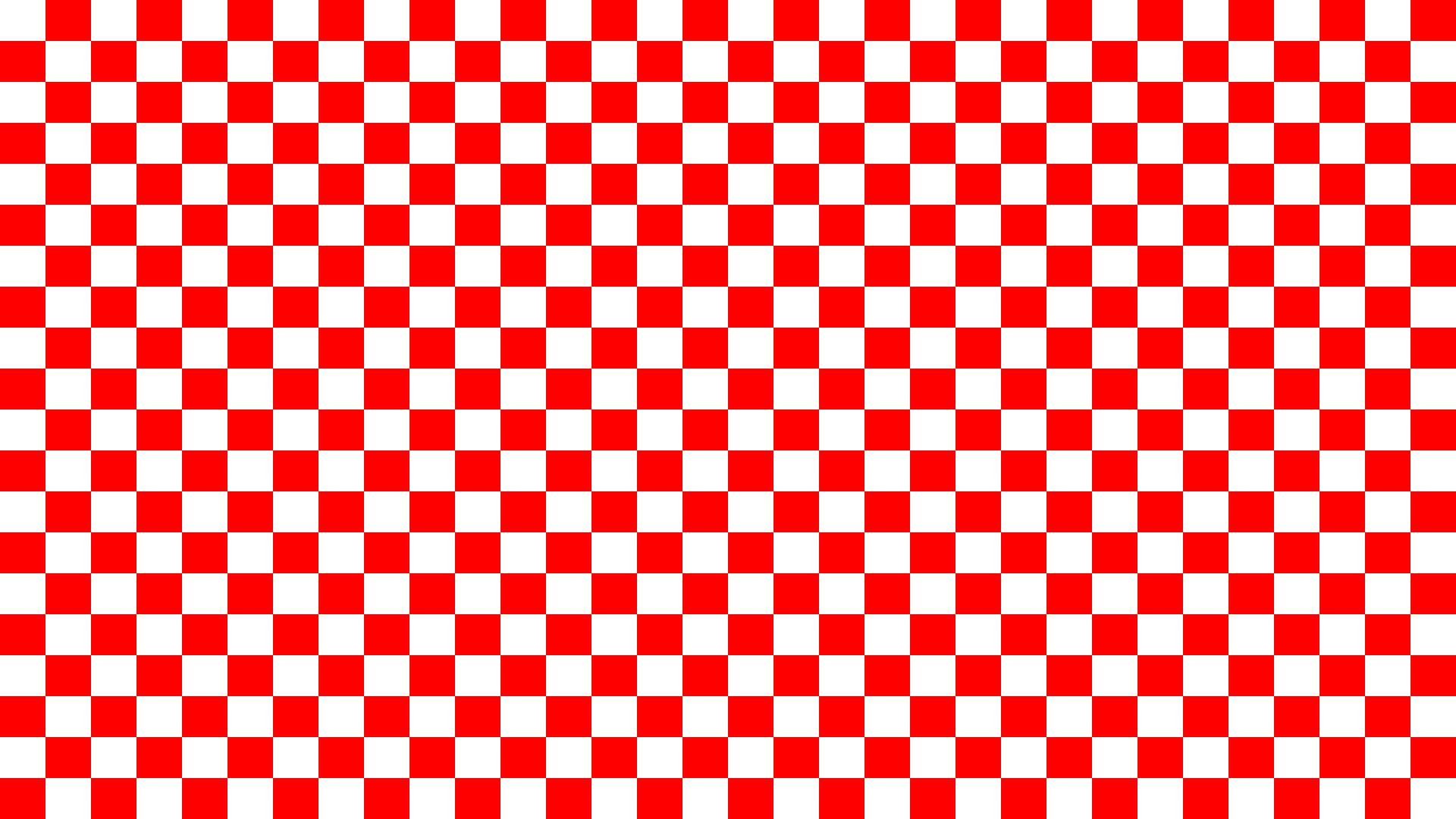} &
\includegraphics[width=0.12\textwidth]{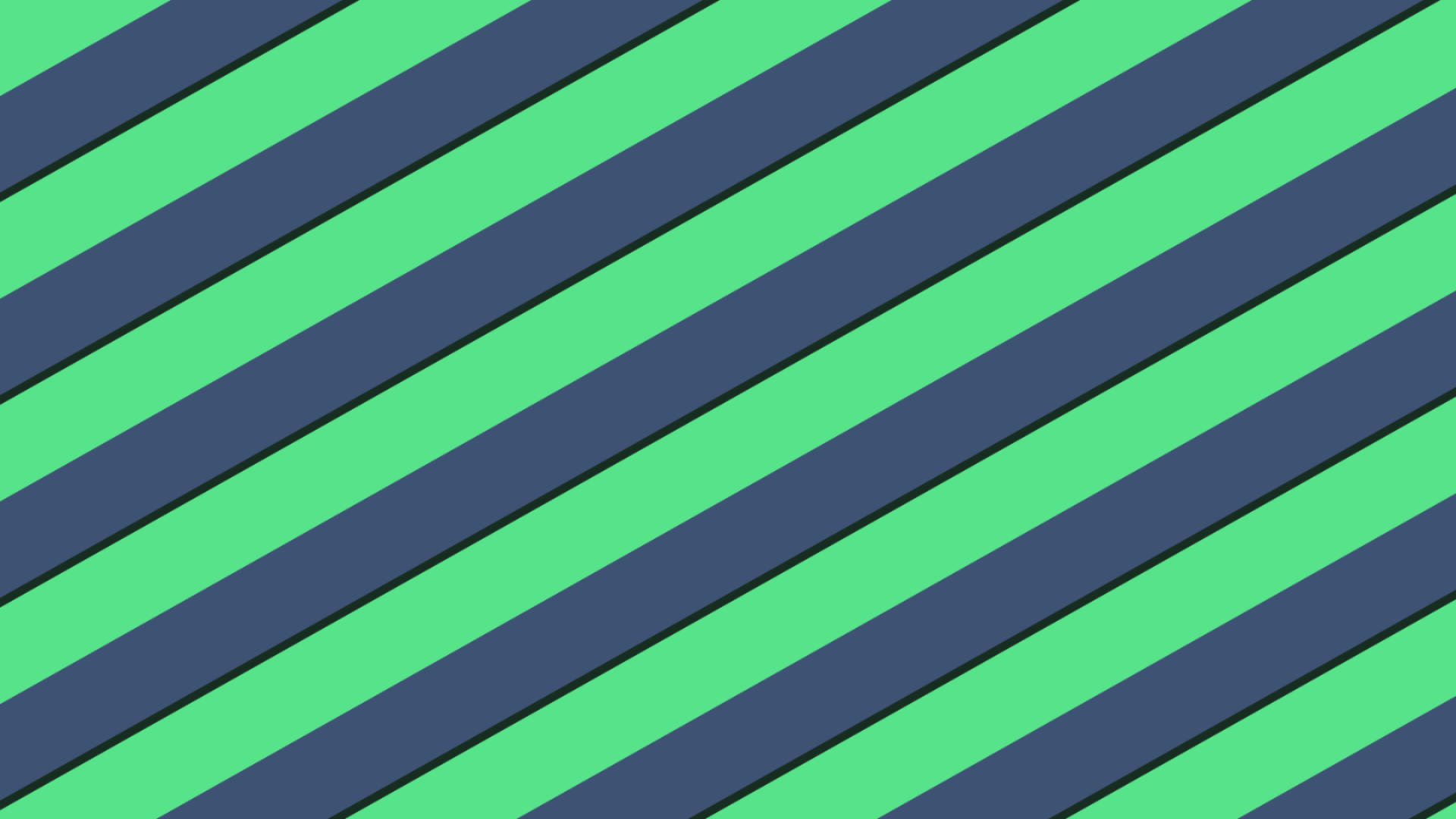} \\
\vspace{-1em}
(a) green-white & (b) red-white & (c) gradient \\[1em]
\end{tabular}
\vspace{-1em}
\caption{Virtual Projector Patterns}
\label{fig:proj_patterns}
\end{figure}

We can observe that the network does learn the desired anisotropic footprint responses very well in general and does not overfit to the training set, but also generalizes to the test set. To relight the object with different virtual projector patterns of (1080, 1920) resolution, we use the following three images shown in Fig. \ref{fig:proj_patterns}.

\begin{figure}[htb]
\centering
\begin{tabular}{cccc}
\includegraphics[width=0.1\textwidth]{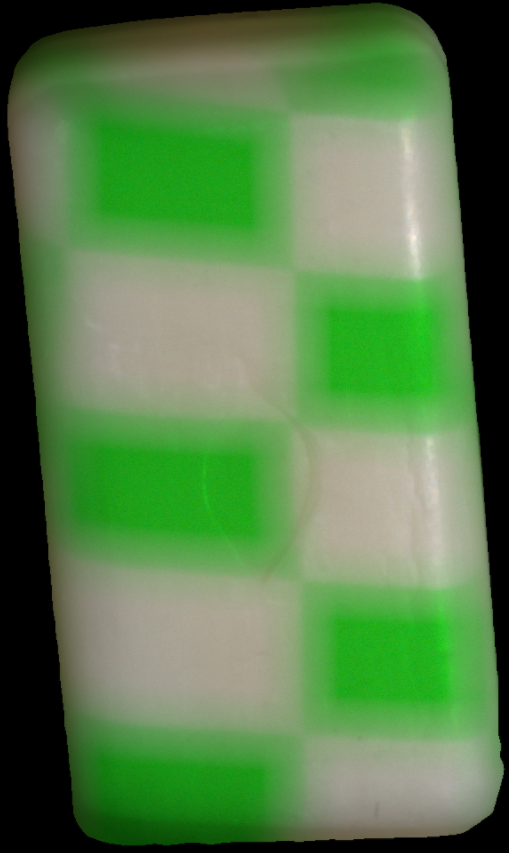} &
\includegraphics[width=0.1\textwidth]{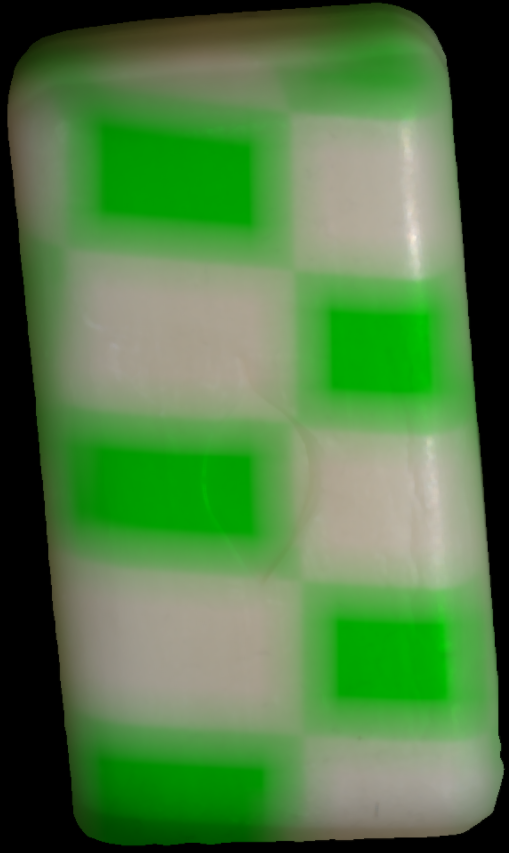} &
\includegraphics[width=0.1\textwidth]{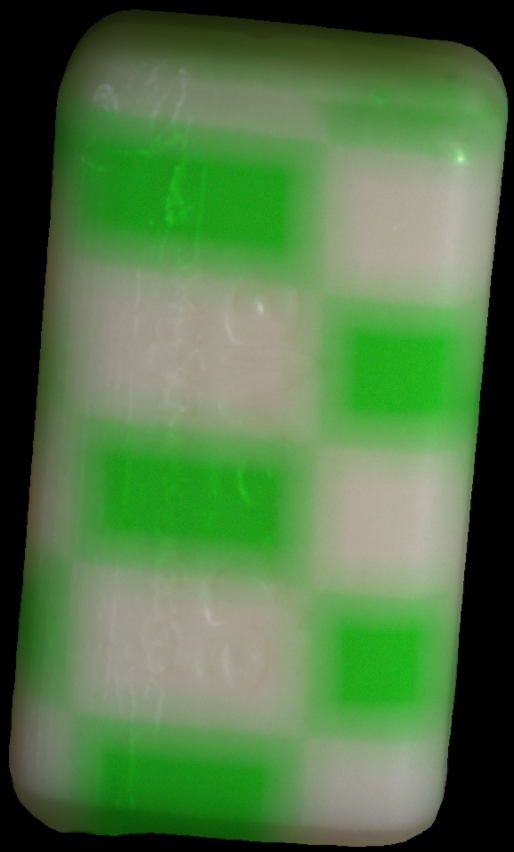} &
\includegraphics[width=0.1\textwidth]{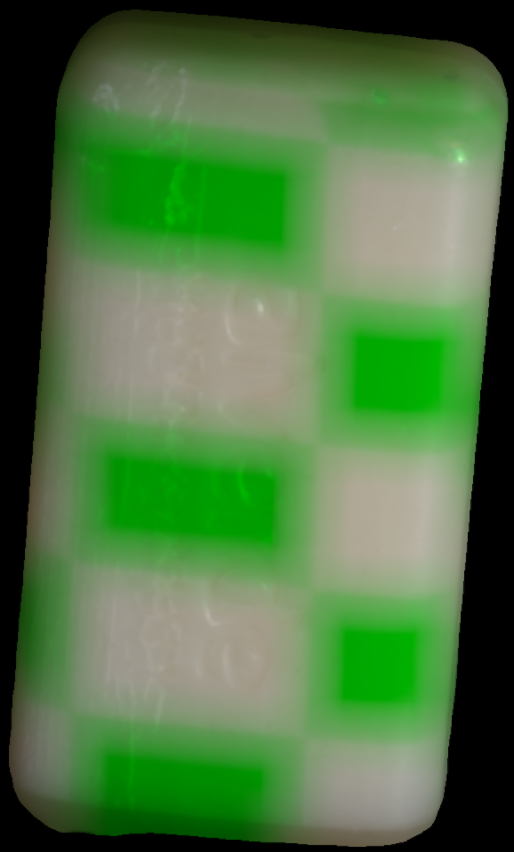} \\
\includegraphics[width=0.1\textwidth]{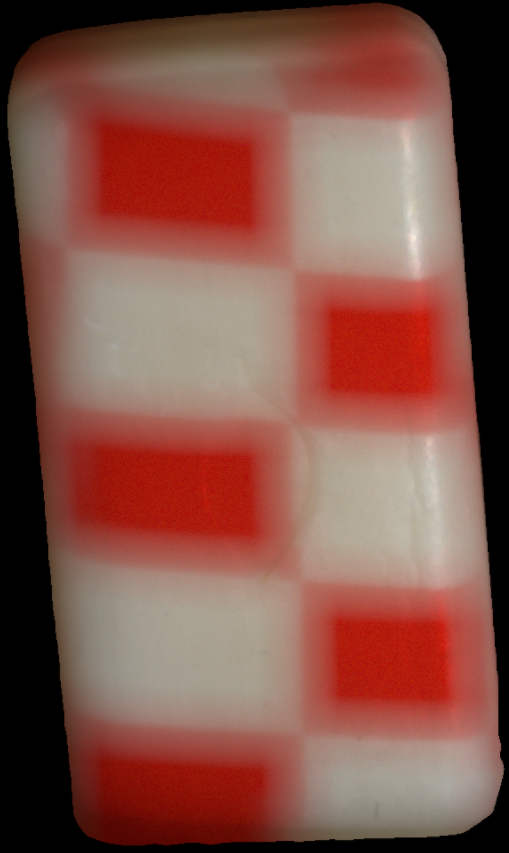} &
\includegraphics[width=0.1\textwidth]{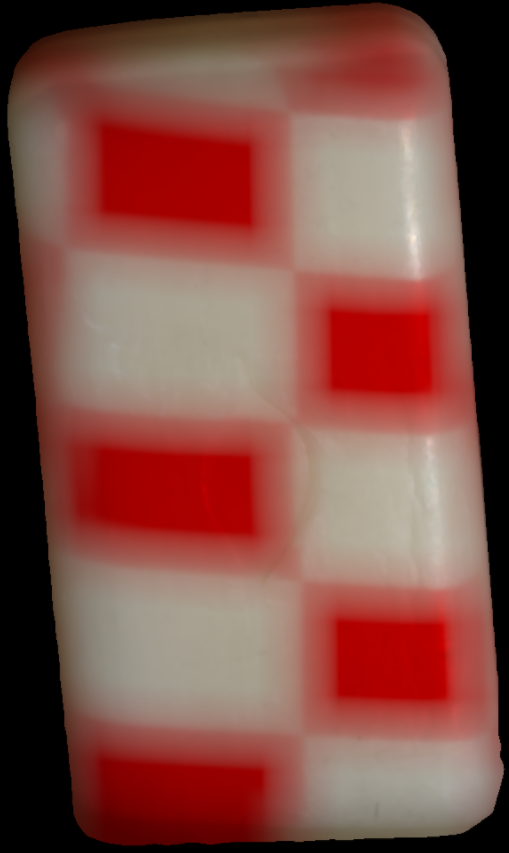} & 
\includegraphics[width=0.1\textwidth]{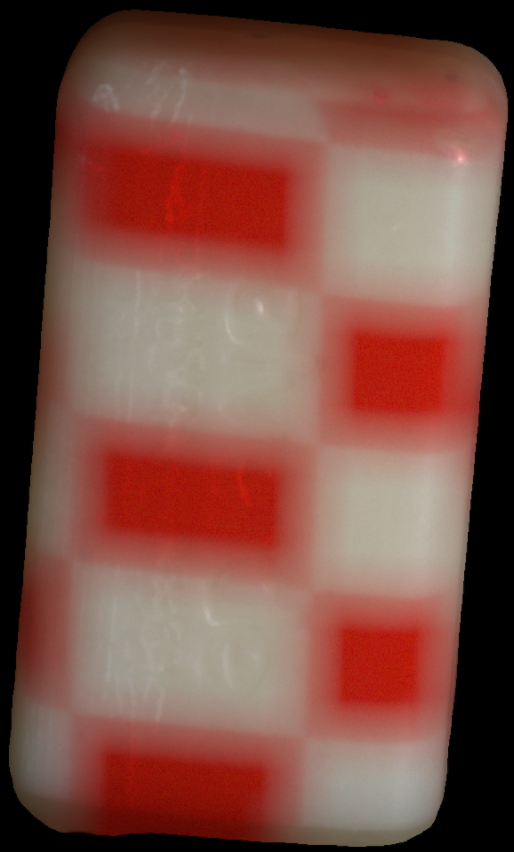} &
\includegraphics[width=0.1\textwidth]{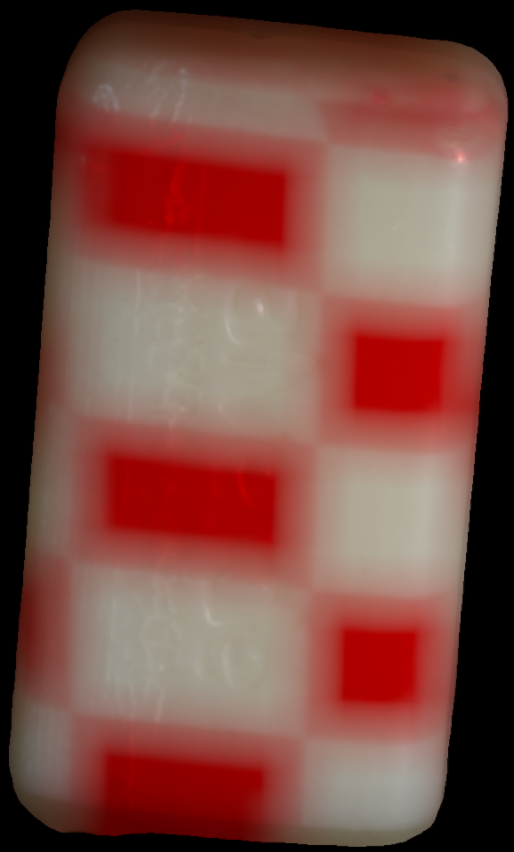} \\

\includegraphics[width=0.1\textwidth]{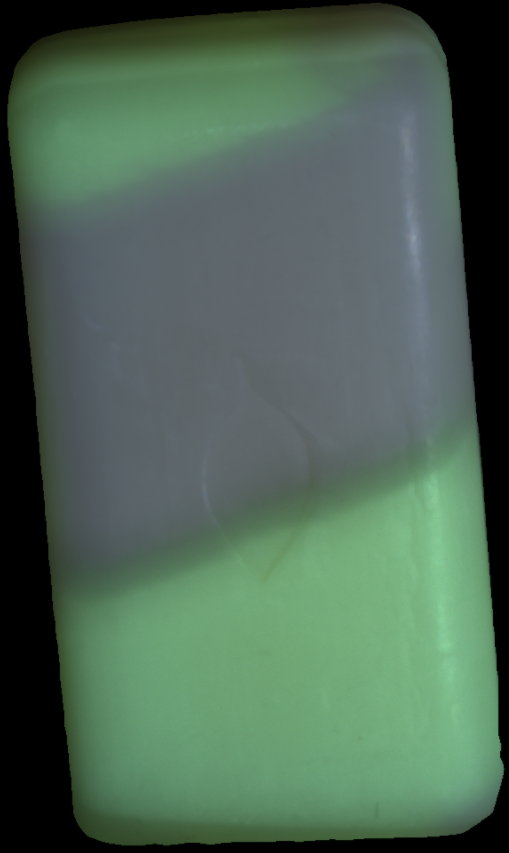} &
\includegraphics[width=0.1\textwidth]{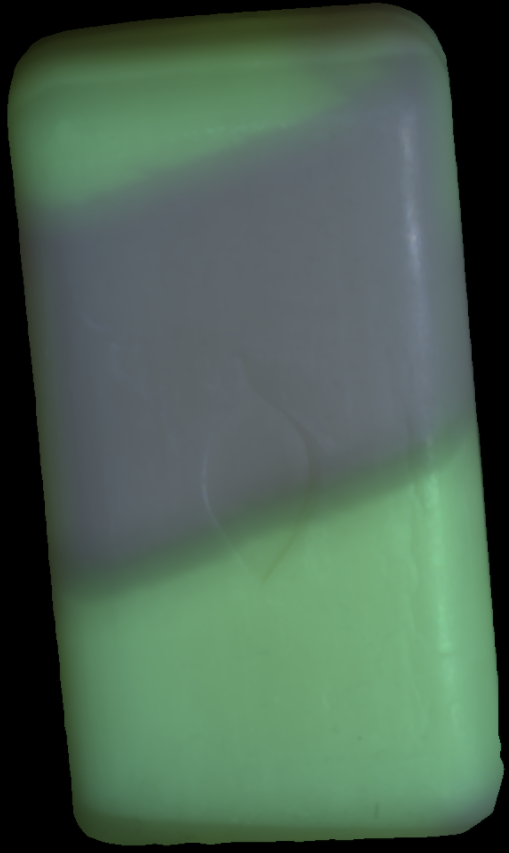} &
\includegraphics[width=0.1\textwidth]{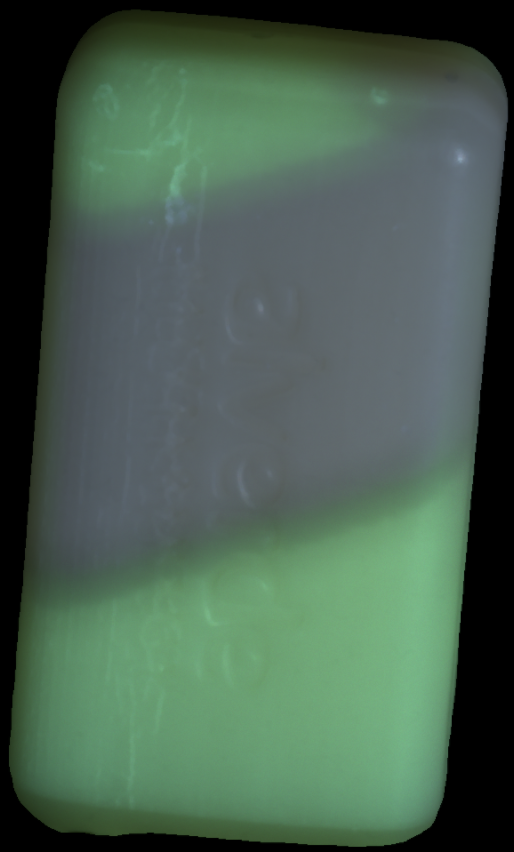} &
\includegraphics[width=0.1\textwidth]{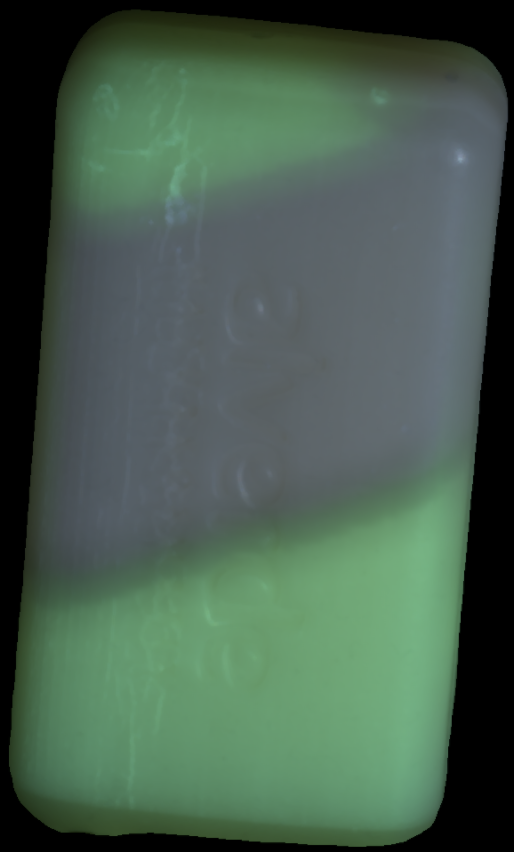}  \\
(a) Target & (b) Output & (c) Target & (d) Output \\[-1em]
\end{tabular}
\caption{Comparison: Target \& Relit Images - Soap front view (first two columns), back view (last two columns). Note how the smooth transition between different colors as well as the indents on the front are accurately reproduced. }
\label{fig:comparison_soap}
\end{figure}

\begin{figure}[htbp]
\centering
\begin{tabular}{cccc}
\includegraphics[width=0.11\textwidth]{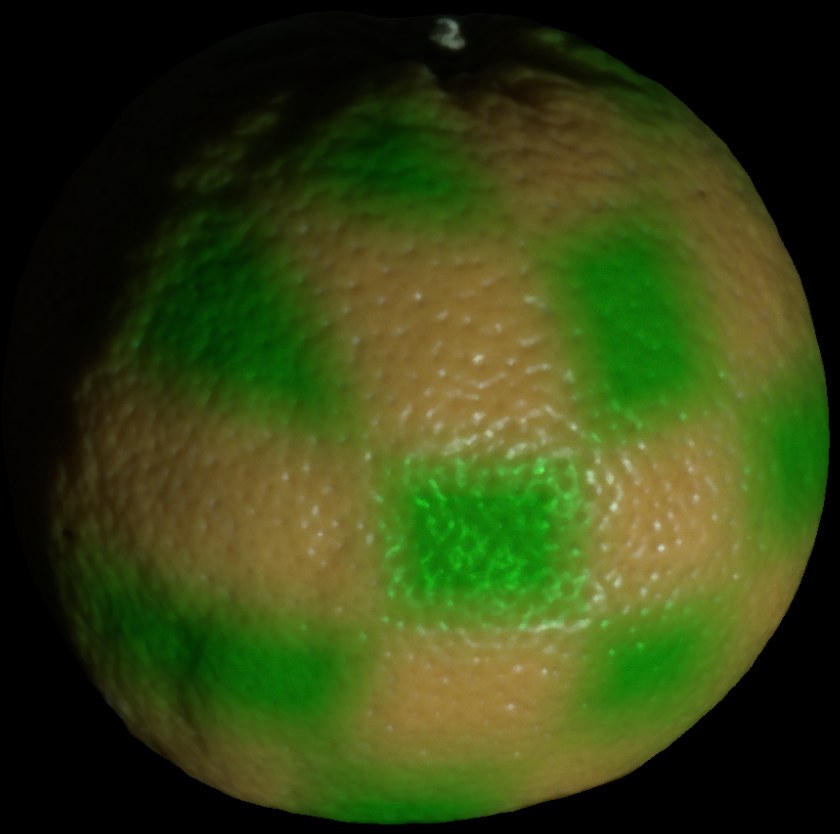} &
\includegraphics[width=0.11\textwidth]{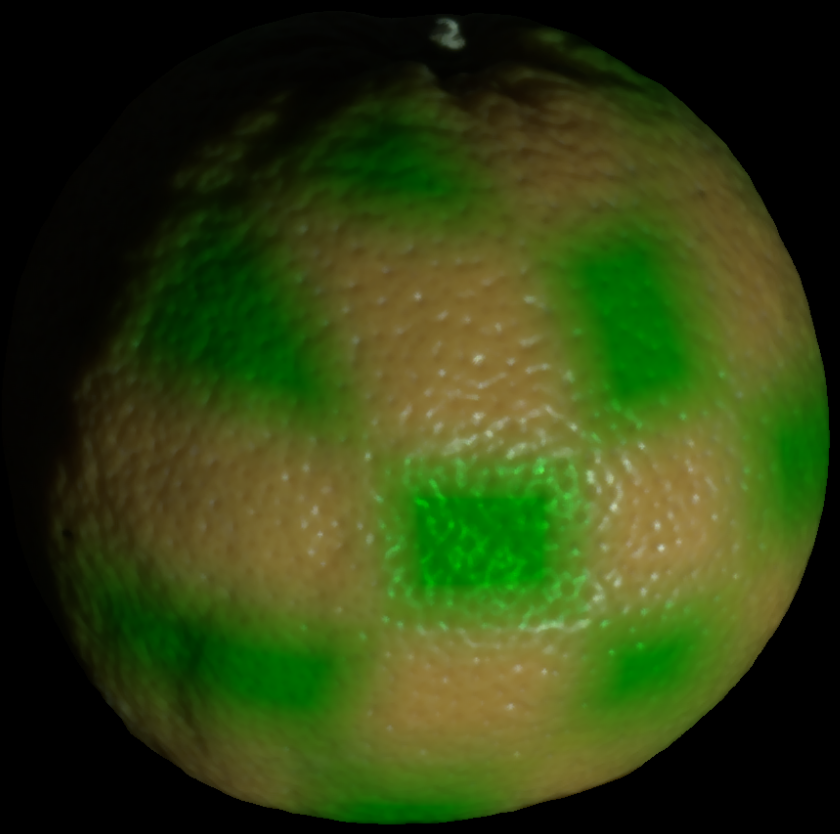} &
\includegraphics[width=0.11\textwidth]{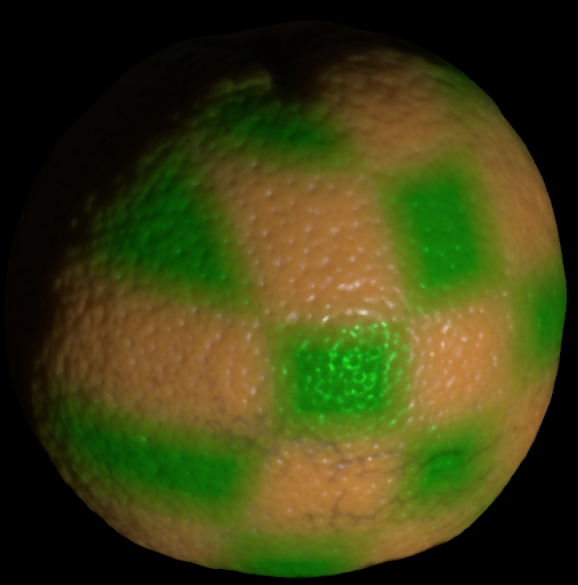} &
\includegraphics[width=0.11\textwidth]{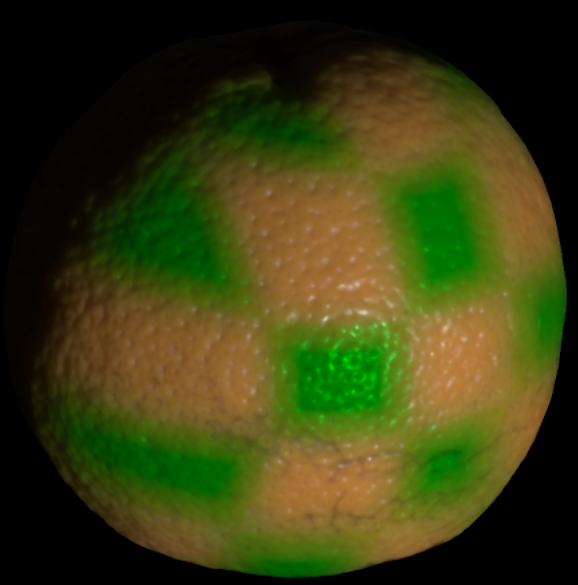} \\
\includegraphics[width=0.11\textwidth]{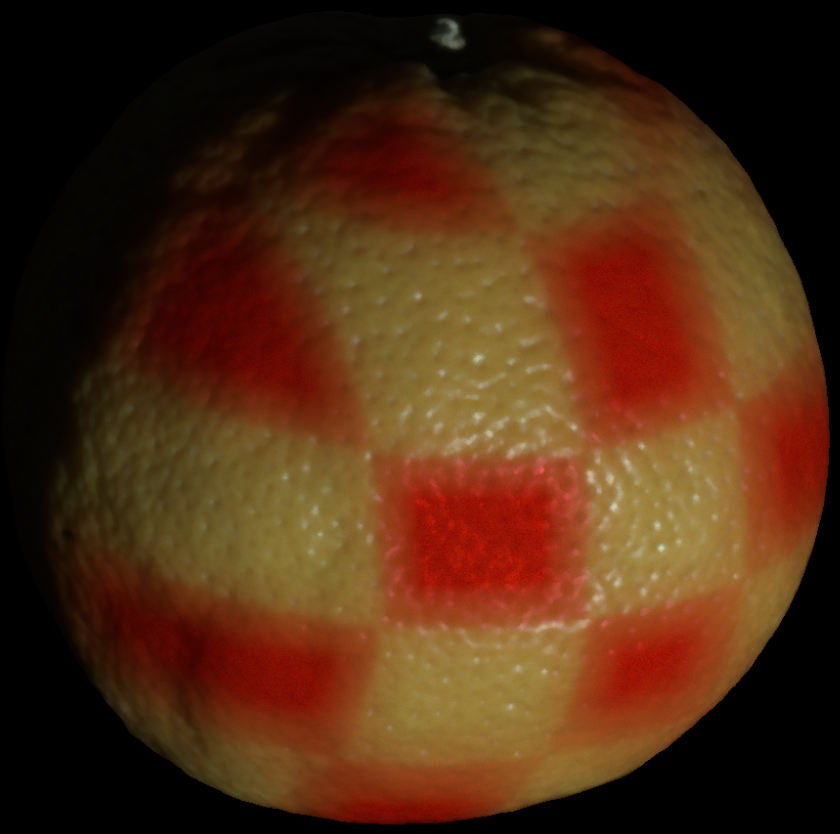} &
\includegraphics[width=0.11\textwidth]{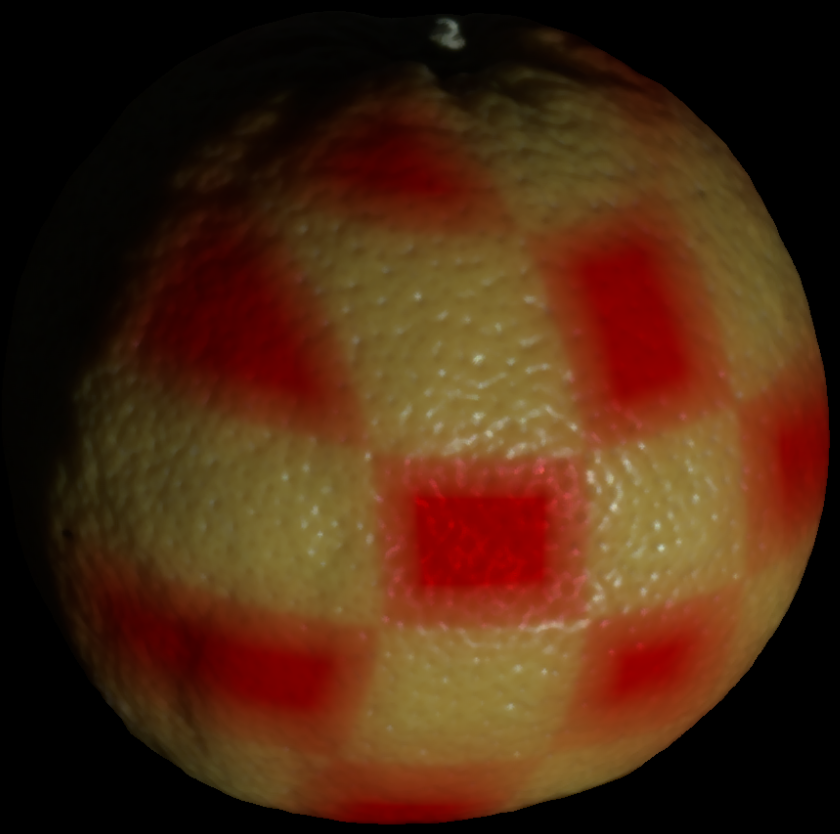} & 
\includegraphics[width=0.11\textwidth]{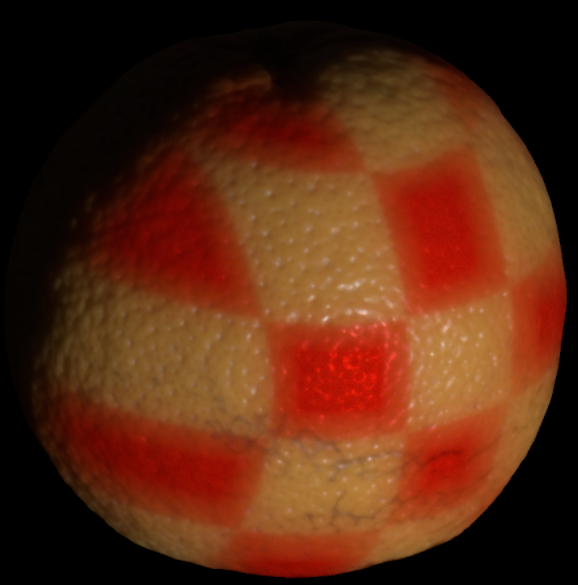} &
\includegraphics[width=0.11\textwidth]{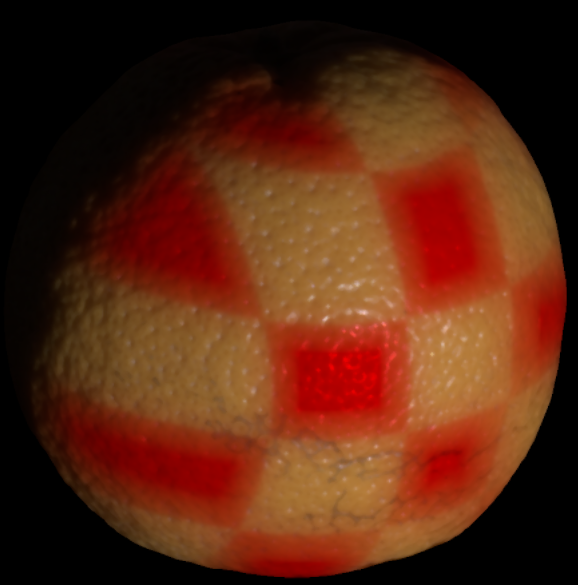} \\
\includegraphics[width=0.11\textwidth]{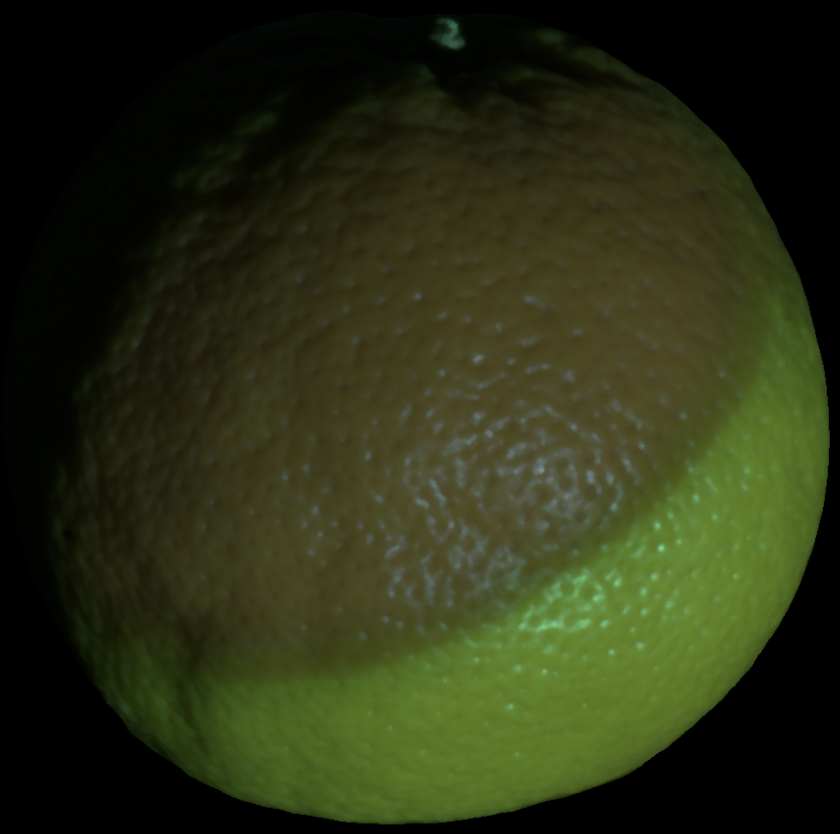} &
\includegraphics[width=0.11\textwidth]{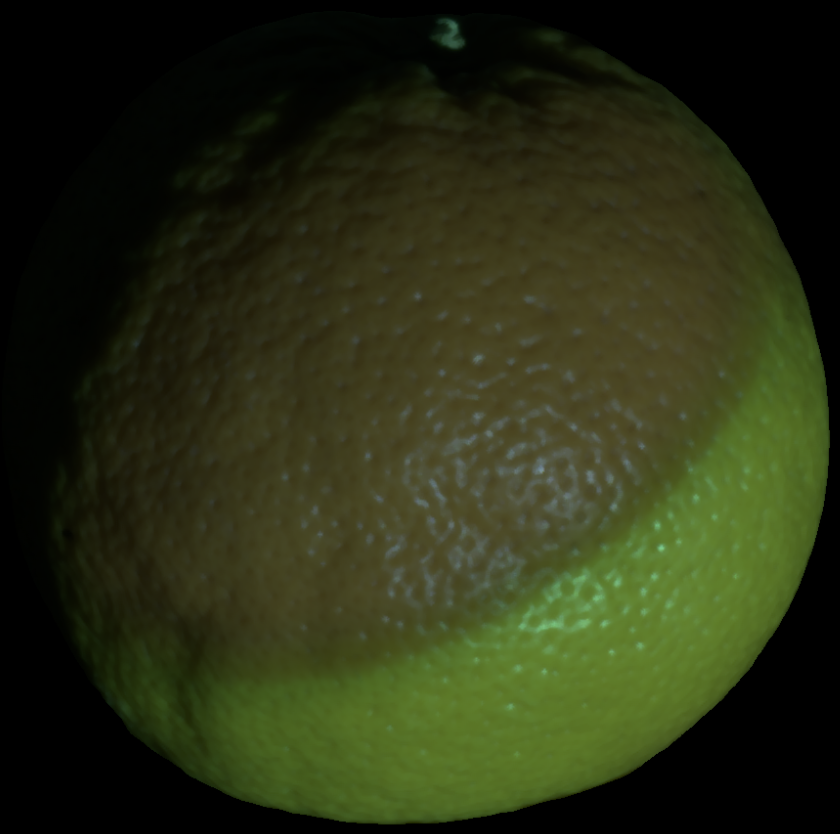} &
\includegraphics[width=0.11\textwidth]{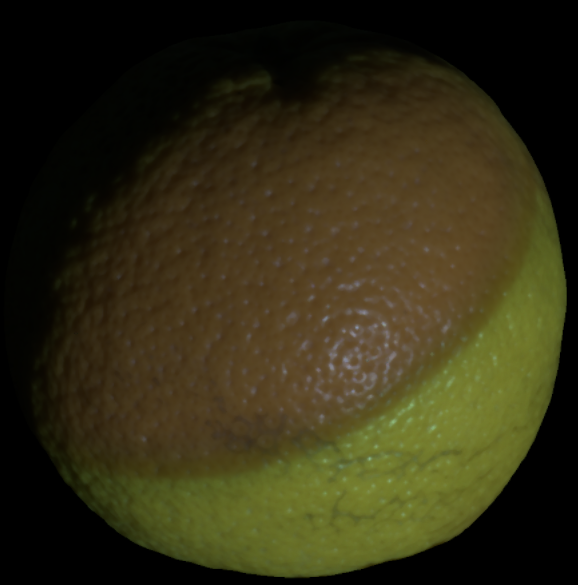} &
\includegraphics[width=0.11\textwidth]{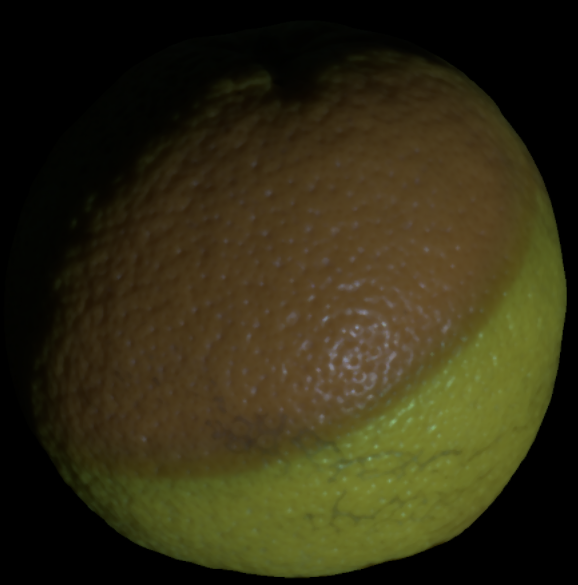} \\
(a) Target & (b) Output & (c) Target & (d) Output \\[-1em]
\end{tabular}
\caption{Comparison: Target \& Relit Images - Orange front view (first two columns), back view (last two columns). Here, the color shift due to the orange peel as well as the peels structure are captured accurately. }
\label{fig:comparison_orange}
\end{figure}

\begin{figure}[htbp]
\centering
\begin{tabular}{cccc}
\includegraphics[width=0.1\textwidth]{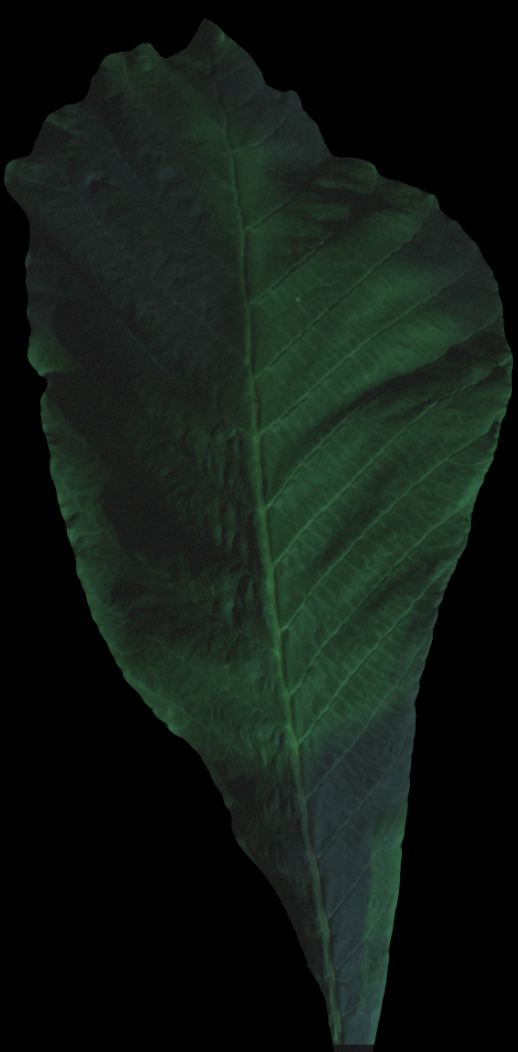} &
\includegraphics[width=0.1\textwidth]{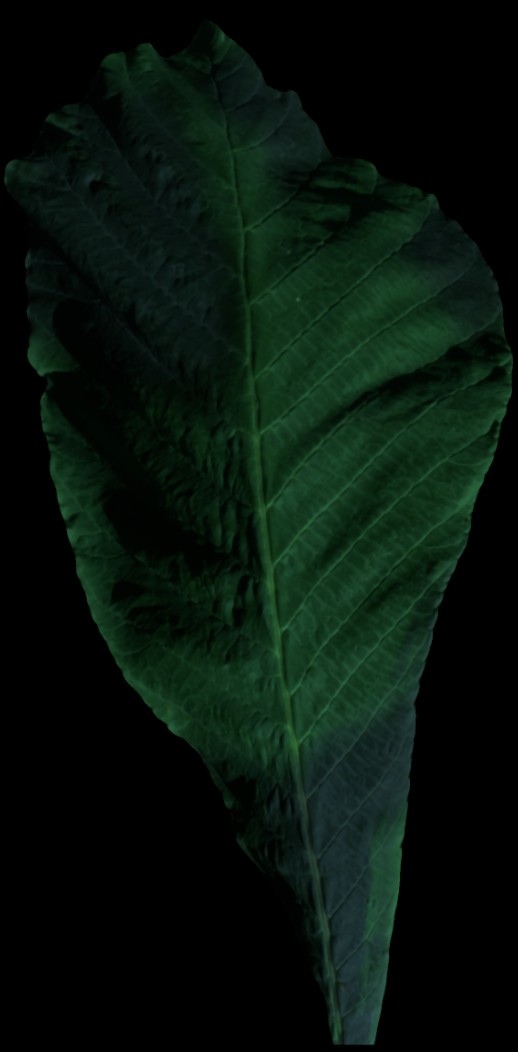} &
\includegraphics[width=0.1\textwidth]{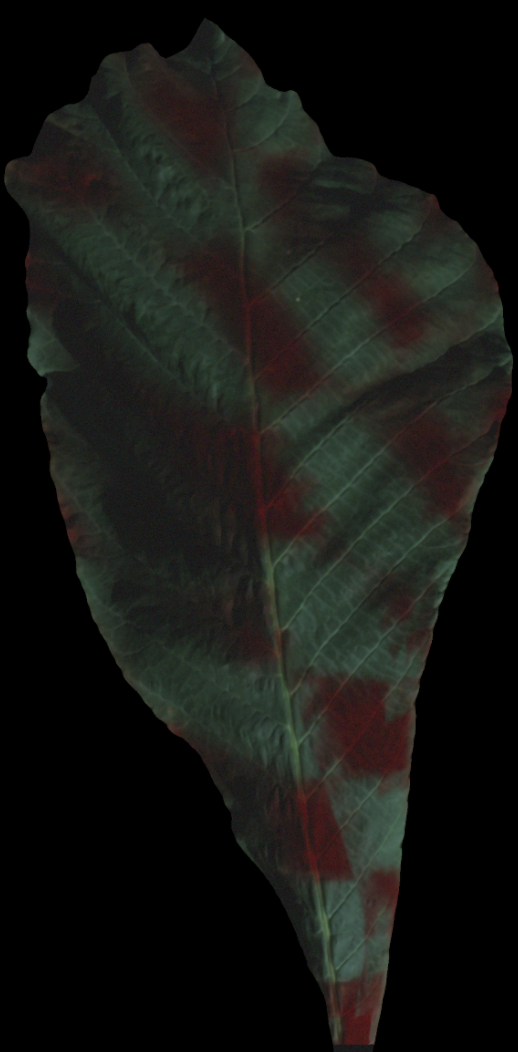} &
\includegraphics[width=0.1\textwidth]{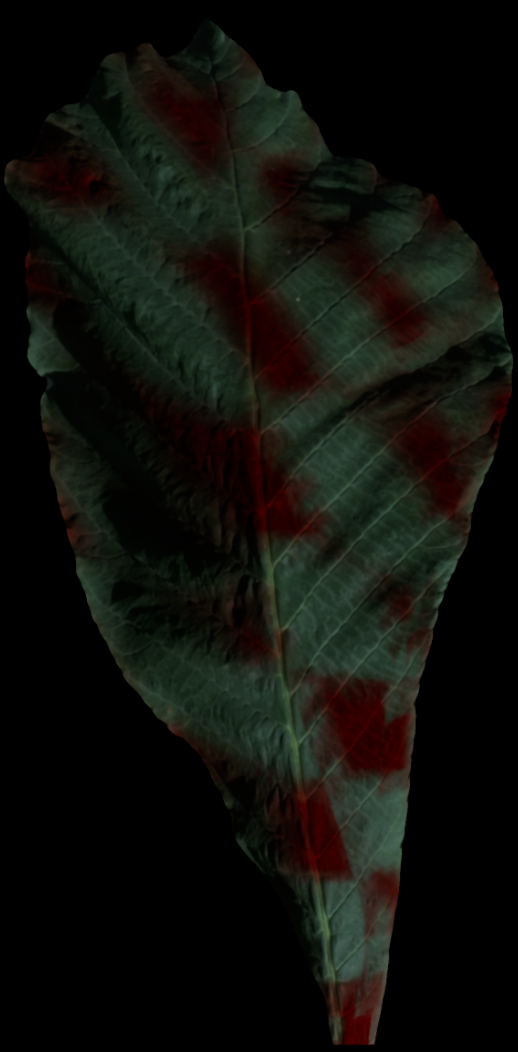} \\
(a) Target & (b) Output & (c) Target & (d) Output \\[-1em]
\end{tabular}
\caption{Comparison: Target \& Relit Images - Leaf front view}
\label{fig:comparison_leaf}
\end{figure}

Fully relit results are presented for a commercially available soap in
Figure~\ref{fig:comparison_soap}, a leaf obtained from a nearby forest in
Figure~\ref{fig:comparison_leaf} and an orange fruit in
Figure~\ref{fig:comparison_orange}. An interesting side note for the captured
leaf is that its rate of decomposition is very fast which means that it starts
to deform in both shape and size just after being plucked from its tree. This
can pose to be a problem if the SAM mask was generated before, while the rest of
the acquisition takes time which might be compounded when acquiring multiple views
of the object. 

Relighting using the predicted footprints after the described color calibration (see Section~\ref{sec:inference}) leads to images that are difficult to distinguish from ground truth. Specifically, the subsurface scattering-induced smooth transition between differently colored patches and the object-dependent color shifts are well captured. In addition, all geometric intricacies are reproduced.
With image based rendering, all material properties, including highlights are
baked in into the footprints. They would not move if one would rotate the
virtual object. 


We also provide quantitative evaluations for the relit results compared  objects captured from multiple views in Table~\ref{tab:image_comparisons}. These numbers underline the very close match between the relit images based on the reconstructed footprints and the ground truth. 

\begin{table}[htbp]
\centering
\small
\begin{tabular}{|l|c|c|c|c|c|}
\hline
\textbf{Object+View} & \textbf{Pattern} & \textbf{MSE$^*$} & \textbf{PSNR} & \textbf{SSIM} & \textbf{LPIPS$^*$} \\
\hline
Soap-front & red-white & $6.14$ & 52.00 & 0.995 & $5.31$ \\
\hline
Soap-front & green-white & $8.42$ & 49.39 & 0.994 & $7.84$ \\
\hline
Soap-front & gradient & $5.75$ & 47.82 & 0.999 & $2.84$ \\
\hline
Soap-back & red-white & $9.52$ & 50.02 & 0.996 & $4.85$ \\
\hline
Soap-back & green-white & $6.22$ & 52.04 & 0.995 & $7.41$ \\
\hline
Soap-back & gradient & $4.70$ & 46.44 & 0.999 & $2.18$ \\
\hline
Orange-front & red-white & $1.62$ & 46.52 & 0.997 & $2.18$ \\
\hline
Orange-front & green-white & $9.5$ & 48.36 & 0.998 & $1.9$ \\
\hline
Orange-front & gradient & $1.0$ & 55.81 & 0.999 & $3.18$ \\
\hline
Orange-back & red-white & $3.6$ & 54.44 & 0.998 & $1.07$ \\
\hline
Orange-back & green-white & $8.56$ & 49.83 & 0.998 & $1.25$ \\
\hline
Orange-back & gradient & $0.8$ & 54.01 & 0.999 & $0.27$ \\
\hline

\end{tabular}
\caption{Quantitative metrics on relit results. $^*$(MSE$\times10^{-6}$, LPIPS$\times10^{-3}$). }
\label{tab:image_comparisons}
\end{table}

\begin{figure}[htbp]
\centering
\begin{tabular}{cccc}
\includegraphics[width=0.1\textwidth]{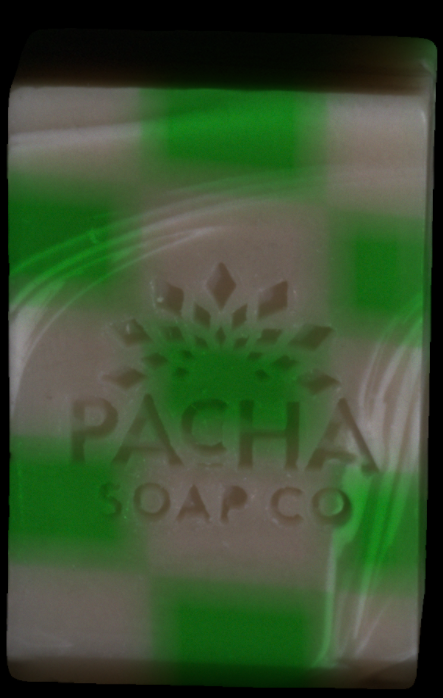} &
\includegraphics[width=0.1\textwidth]{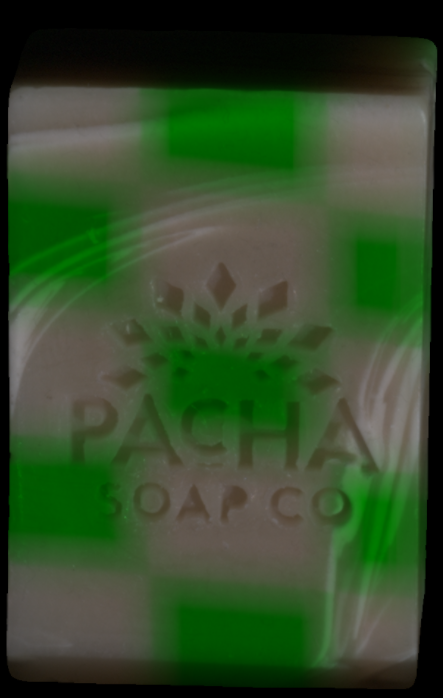} &
\includegraphics[width=0.1\textwidth]{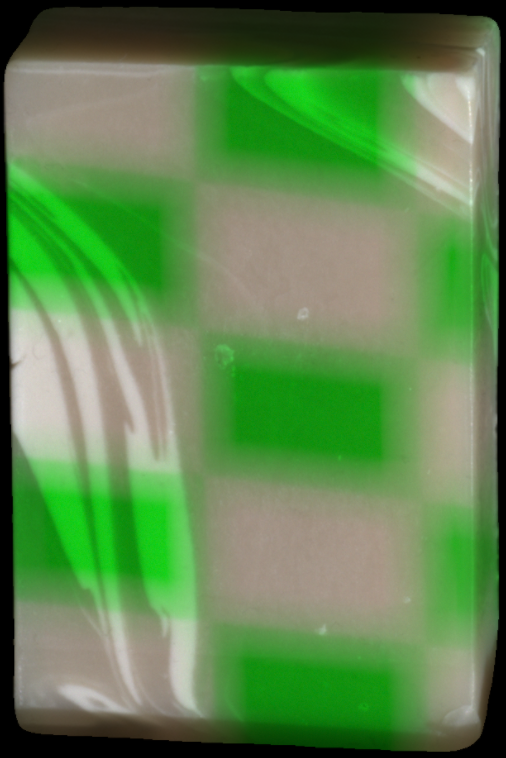} &
\includegraphics[width=0.1\textwidth]{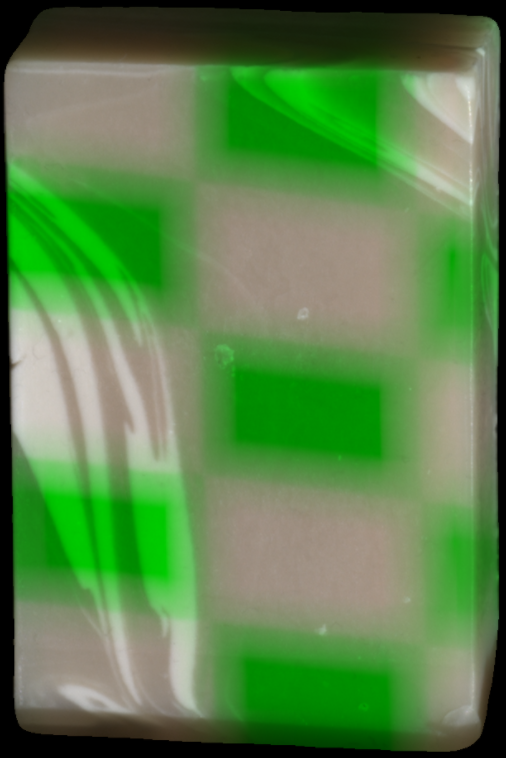} \\
\includegraphics[width=0.1\textwidth]{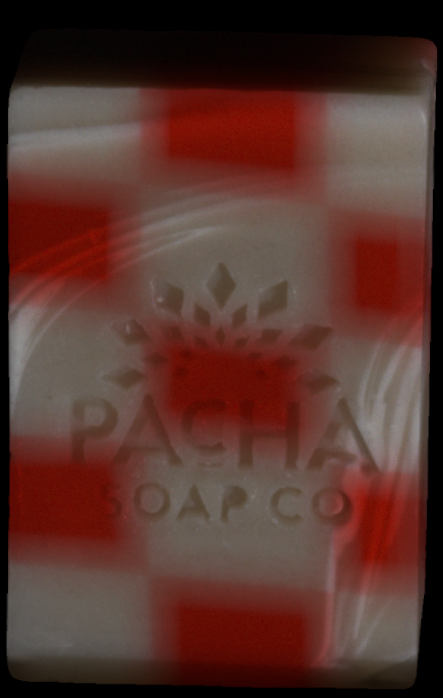} &
\includegraphics[width=0.1\textwidth]{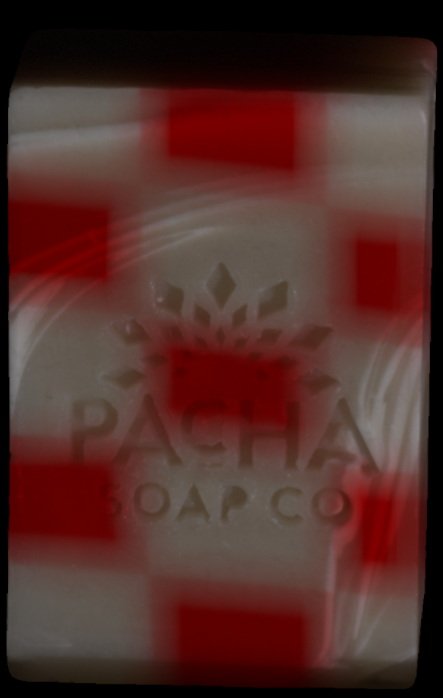} &
\includegraphics[width=0.1\textwidth]{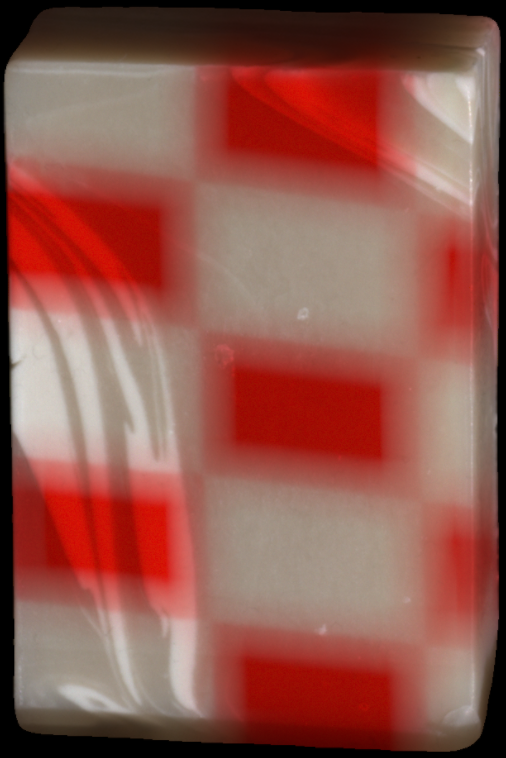} &
\includegraphics[width=0.1\textwidth]{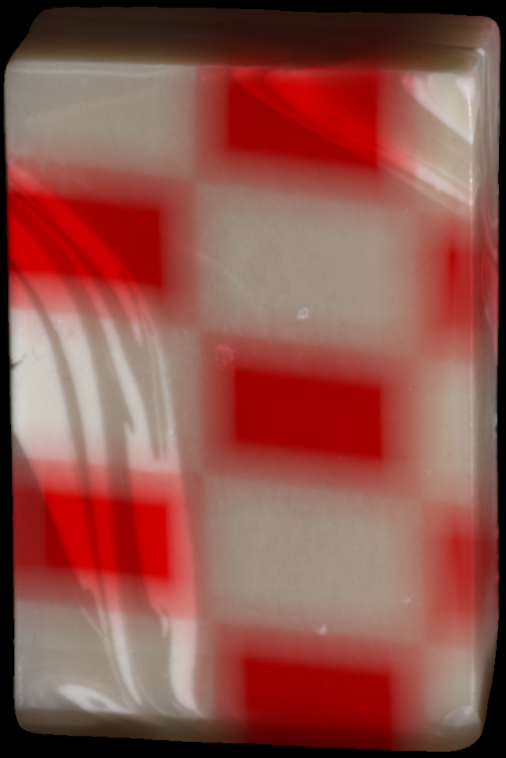} \\
\includegraphics[width=0.1\textwidth]{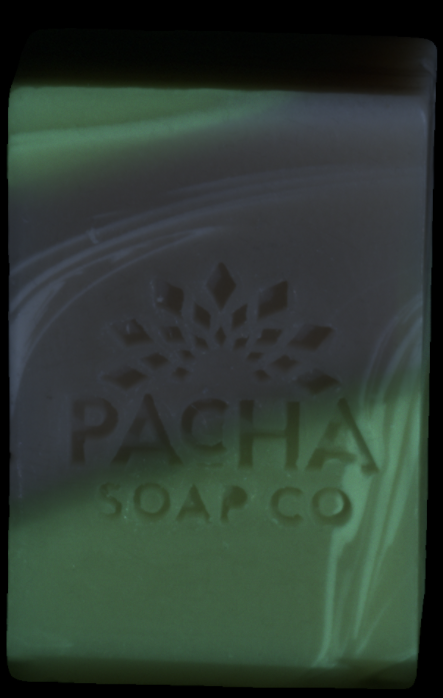} &
\includegraphics[width=0.1\textwidth]{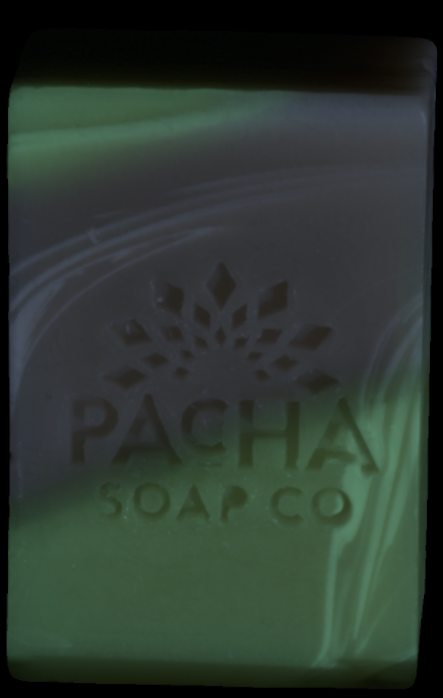} &
\includegraphics[width=0.1\textwidth]{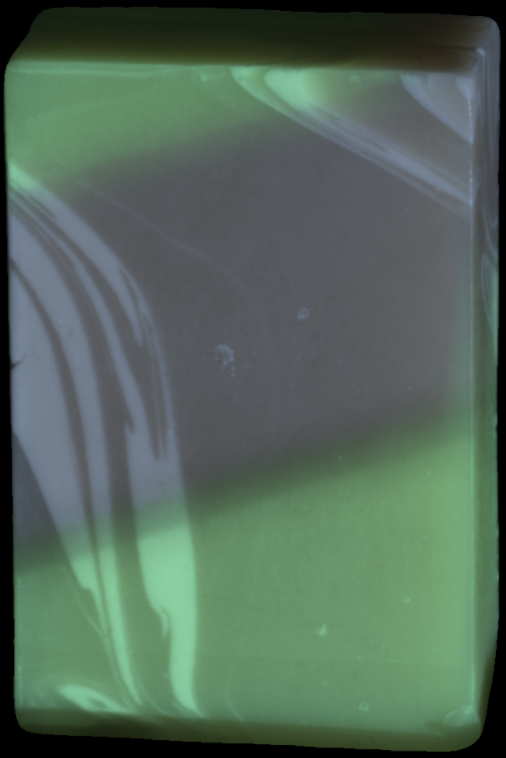} &
\includegraphics[width=0.1\textwidth]{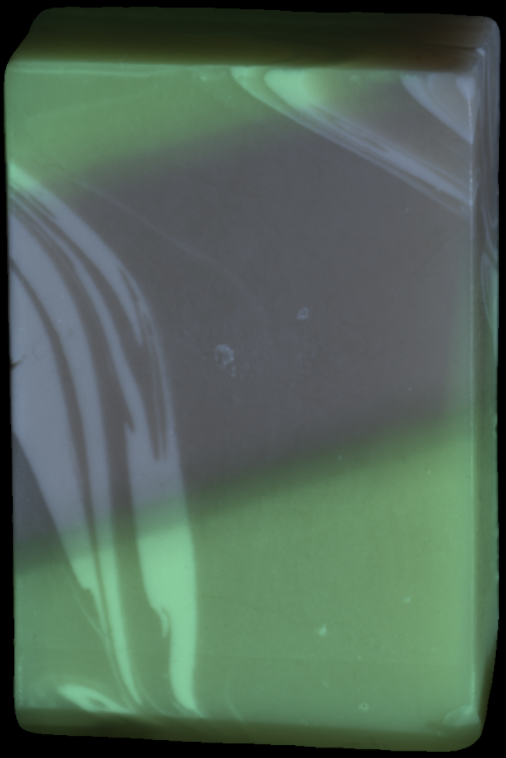} \\
(a) Target & (b) Output & (c) Target & (d) Output \\[-1em]
\end{tabular}
\caption{Comparison: Target \& Relit Images - \textbf{Unseen} Soap2 front \&
back view. No images of this object had been used during training.}
\label{fig:comparison_soap_pacha_frontview}
\end{figure}

Finally, we use the pre-trained attention U-Net, which has been trained on the
training objects from multiple views for an unseen soap (Soap2) also for
multiple views (see Figure~\ref{fig:comparison_soap_pacha_frontview} and
supplemental materials). Even though the U-Net has never seen any input of the
second soap during training, all object-specific features, such as the
appearance of the imprint as well as the brighter veins and the scattering
around them are perfectly reconstructed. 
The corresponding metrics are listed in Table~\ref{tab:unseen_object_image_comparisons}.

We use a frozen U-Net pre-trained on Soap, Orange and Leaf to relite a
geometrically challenging 3D printed object to show that our method generalizes
to such objects as well, which is provided in the supplementary.


\begin{table}[htbp]
\centering
\small
\begin{tabular}{|l|c|c|c|c|c|}
\hline
\textbf{Object+View} & \textbf{Pattern} & \textbf{MSE$^*$} & \textbf{PSNR} & \textbf{SSIM} & \textbf{LPIPS$^*$} \\
\hline
Soap-front & red-white & $1.45$ & 37.54 & 0.997 & $3.9$ \\
\hline
Soap-front & green-white & $1.37$ & 37.28 & 0.995 & $5.0$ \\
\hline
Soap-front & gradient & $0.29$ & 43.04 & 0.999 & $1.3$ \\
\hline
Soap-back & red-white & $1.28$ & 48.18 & 0.998 & $1.1$ \\
\hline
Soap-back & green-white & $1.22$ & 48.09 & 0.997 & $1.5$ \\
\hline
Soap-back & gradient & $0.22$ & 52.24 & 0.999 & $0.33$ \\
\hline

\end{tabular}
\caption{Quantitative metrics on relit images for the unseen object Soap2. $^*$(MSE$\times10^{-5}$, LPIPS$\times10^{-3}$). }
\label{tab:unseen_object_image_comparisons}
\end{table}

\section{Conclusions}

To obtain a relightable representation for subsurface scattering objects, we
present a novel technique that can predict the per-pixel spatially varying pixel
footprints from just six input images captured with high-frequency horizontal
and vertical sinusoidal patterns. A trained U-Net translates those phase-shifted
profilometry patterns, which at the same time can be used for 3D scanning. The
training of the U-Net requires a paired dataset of sinusoidal patterns with
corresponding impulse response scattering footprints. But those could be
acquired for a completely different view or on different objects, while the CNN
generalized well to novel views and novel objects. The predicted scattering
functions accurately model the actual scattering behaviour, resulting in relit
images that are hard to distinguish from real-world captured objects. 
While our approach allows for subsurface scattering representations from very few measurements, in future work, a couple of further effects will need to be addressed: This should cover the separation of surface reflections and subsurface scattering, scattering effects that extend beyond the current size of the estimated footprint, as well as angular dependencies.

\section{Acknowledgement}
The work described in this paper was funded by: Sony Europe Limited, Germany. The authors thank Prasan Ashok Shedligeri, Zoltan Facius and Alexander Gatto for their support. The author would also like to extend their gratitude and thanks to the  International Max Planck Research School for Intelligent Systems (IMPRS-IS) for supporting Arjun Majumdar.

\bibliographystyle{eg-alpha-doi} 
\bibliography{egbibsample}       



\end{document}